\PassOptionsToPackage{warn}{textcomp}
\documentclass[sigconf]{acmart}
\usepackage{booktabs}
\usepackage{times}
\usepackage{epsfig}
\usepackage{graphicx}
\usepackage{amsmath}

\usepackage{amssymb}
\usepackage{times}
\usepackage{soul}
\usepackage{url}
\usepackage[utf8]{inputenc}
\usepackage{amsmath}
\usepackage{amsthm}
\usepackage{booktabs}
\usepackage{subfigure}
\usepackage{amsmath}
\usepackage{amssymb}
\usepackage{booktabs}
\usepackage{multirow}
\usepackage[american]{babel}
\usepackage{microtype}
\usepackage{lipsum}
\usepackage{bm}
\usepackage[noend]{algpseudocode}
\usepackage{paralist,algorithm}
\usepackage{balance}

\newcommand{\x}{{\bf x}}

\newcommand{\D}{\mathcal{D}}
\newcommand{\T}{\mathcal{T}}
\newcommand{\Y}{\mathcal{Y}}
\newcommand{\R}{\mathbb{R}}

\newcommand{\eg}{\emph{e.g.}}
\newcommand{\ie}{\emph{i.e.}}

\newcommand{\bfname}[1]{{\bf #1}}
\newcommand{\name}{{\sc Coil }}
\newcommand{\mame}{{\sc Coil}}

\AtBeginDocument{%
  \providecommand\BibTeX{{%
    \normalfont B\kern-0.5em{\scshape i\kern-0.25em b}\kern-0.8em\TeX}}}

\setcopyright{acmcopyright}

\copyrightyear{2021}
\acmYear{2021}
\setcopyright{acmcopyright}\acmConference[MM '21]{Proceedings of the 29th ACM International Conference on Multimedia}{October 20--24, 2021}{Virtual Event, China}
\acmBooktitle{Proceedings of the 29th ACM International Conference on Multimedia (MM '21), October 20--24, 2021, Virtual Event, China}
\acmPrice{15.00}
\acmDOI{10.1145/3474085.3475306}
\acmISBN{978-1-4503-8651-7/21/10}

\begin{document}

\title{Co-Transport for Class-Incremental Learning}

\author{Da-Wei Zhou, Han-Jia Ye$^\dagger$, De-Chuan Zhan}
\affiliation{%
  \institution{State Key Laboratory for Novel Software Technology, Nanjing University}
  }
\email{{zhoudw, yehj}@lamda.nju.edu.cn, zhandc@nju.edu.cn}

\begin{abstract}

	Traditional learning systems are trained in closed-world for a fixed number of classes, and need pre-collected datasets in advance. 
	However, new classes often emerge in real-world applications and should be learned incrementally. For example, in electronic commerce, new types of products appear daily, and in a social media community, new topics emerge frequently.
	Under such circumstances, incremental models should learn several new classes at a time without forgetting.
	We find a strong correlation between old and new classes in incremental learning, which can be applied to relate and facilitate different learning stages mutually.
	As a result, we propose CO-transport for class Incremental Learning (\mame), which learns to relate across incremental tasks with the class-wise semantic relationship. 
	In detail, co-transport has two aspects: \emph{prospective transport} tries to augment the old classifier with optimal transported knowledge as fast model adaptation. 
	\emph{Retrospective transport} aims to transport new class classifiers backward as old ones to overcome  forgetting. 
	With these transports,  \name efficiently adapts to new tasks, and stably resists forgetting.
	Experiments on benchmark and real-world multimedia datasets validate the effectiveness of our proposed method.
\end{abstract}



\maketitle
\footnotetext[1]{Correspondence to: Han-Jia Ye (yehj@lamda.nju.edu.cn)}

\section{Introduction}
\begin{figure}[t]
	\begin{center}
		\includegraphics[width=1\columnwidth]{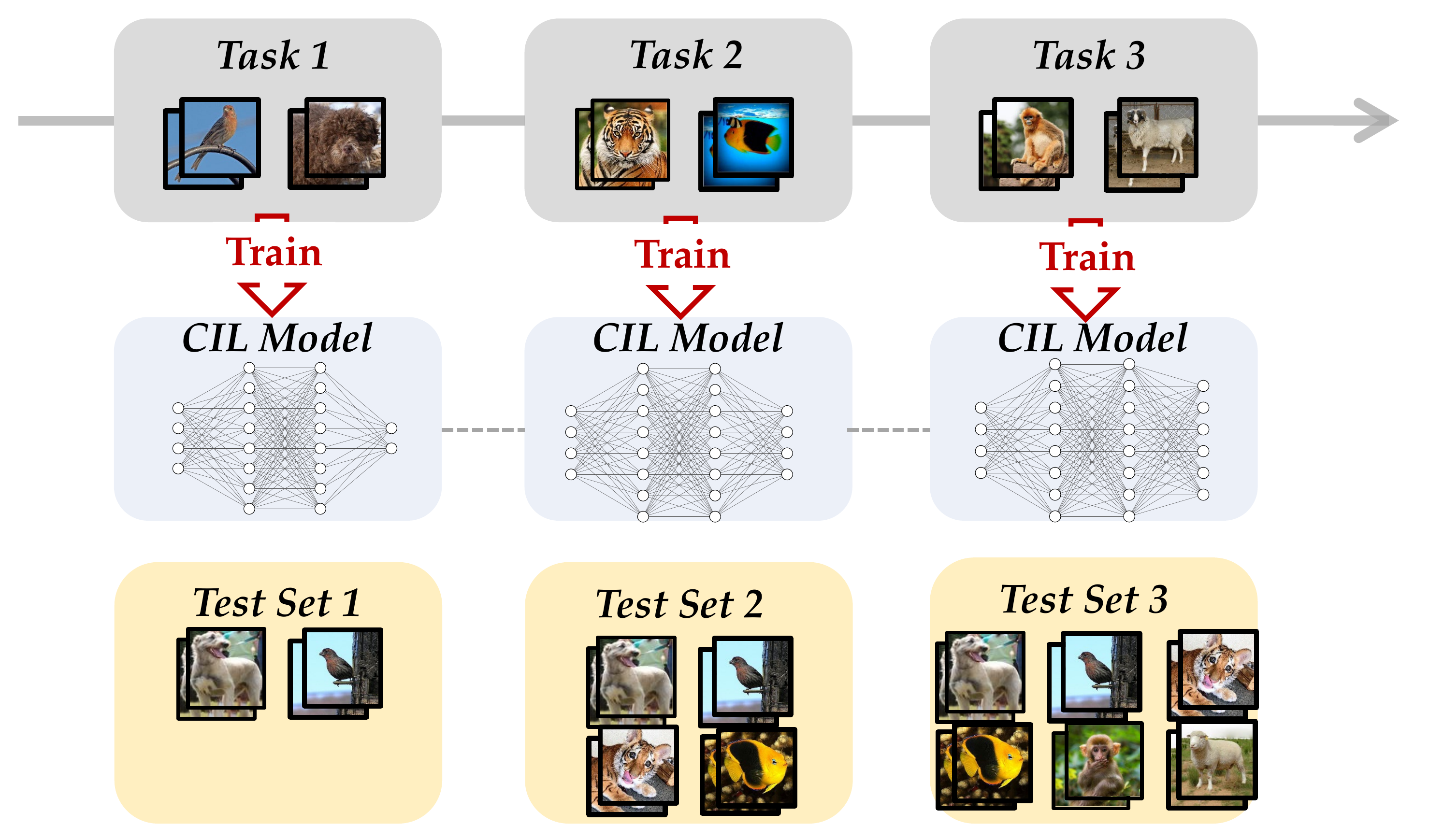}
	\end{center}
	\vspace{-3mm}
	\caption{\small  The setting of class-incremental learning. Non-overlapping classes arrive sequentially, and we need to build a classifier for all the classes incrementally. We only have access to the classes of the current task, and need to learn the new classes with the current model. After the learning stage of each task, the model is evaluated among all seen classes, \ie, it should not only perform well in the newly learned classes, but also remember the former  classes without catastrophic forgetting.
	} \label{figure:incremental}
\end{figure}

With the development of deep learning, current deep models can learn a fixed number of classes with high performance. However, in our ever-changing world, data often comes from an open environment, which is  with stream format~\cite{golab2003issues} or available temporarily due to privacy issues~\cite{de2019continual}.
To tackle this, the classifier should learn new classes incrementally instead of restarting the training process~\cite{li2017learning}. A straightforward approach is to finetune the model on the incoming new data, while it suffers \emph{catastrophic forgetting} phenomena~\cite{mccloskey1989catastrophic}: due to the absence of previous data, the prediction on former classes drastically drops. Class-incremental learning (CIL)~\cite{rebuffi2017icarl} aims to extend the acquired knowledge with only new classes. 
For example, in online opinion
monitoring, new topics often emerge as news happens~\cite{masud2010}, and in the electronic commerce platform, new types of products appear daily~\cite{luo2020alicoco,xu2019open}. 
In a real-world face recognition system, the face classes are increasing as time goes, and the model needs to learn to classify more classes incrementally~\cite{zheng2020learning}. 
Figure~\ref{figure:incremental} demonstrates the setting of CIL. In the first task, the model needs to classify  birds and dogs. After that, the model is incrementally updated with two new classes, \ie, tigers and fish, and it needs to classify among two old classes (birds and dogs) and two new classes (tigers and fish). New categories arrive progressively, and the model needs to classify more classes without forgetting former ones.

According to whether saving old class instances, CIL algorithms can be divided into two groups.
Without saving any instances, \cite{kirkpatrick2017overcoming,zenke2017continual,aljundi2018memory,lee2017overcoming} regularize important parameters from drastically changing to prevent forgetting. The differences in them lie in the way of parameter importance calculation, \eg, by Fisher Information Matrix or by loss-based importance weight estimations.
Other works selectively save exemplars from old classes and rehearsal them when learning new tasks~\cite{rebuffi2017icarl,IscenZLS20,xiang2019incremental}. 
In incremental learning, the model should have transferability, \ie, the former learned knowledge should decrease the difficulty when learning new classes. Correspondingly, the newly learned classes should consolidate former knowledge.  However, these approaches only use the old model to prevent forgetting but ignore  facilitating new classes learning process. 

\begin{figure}[t]
	\begin{center}
		\includegraphics[width=1\columnwidth]{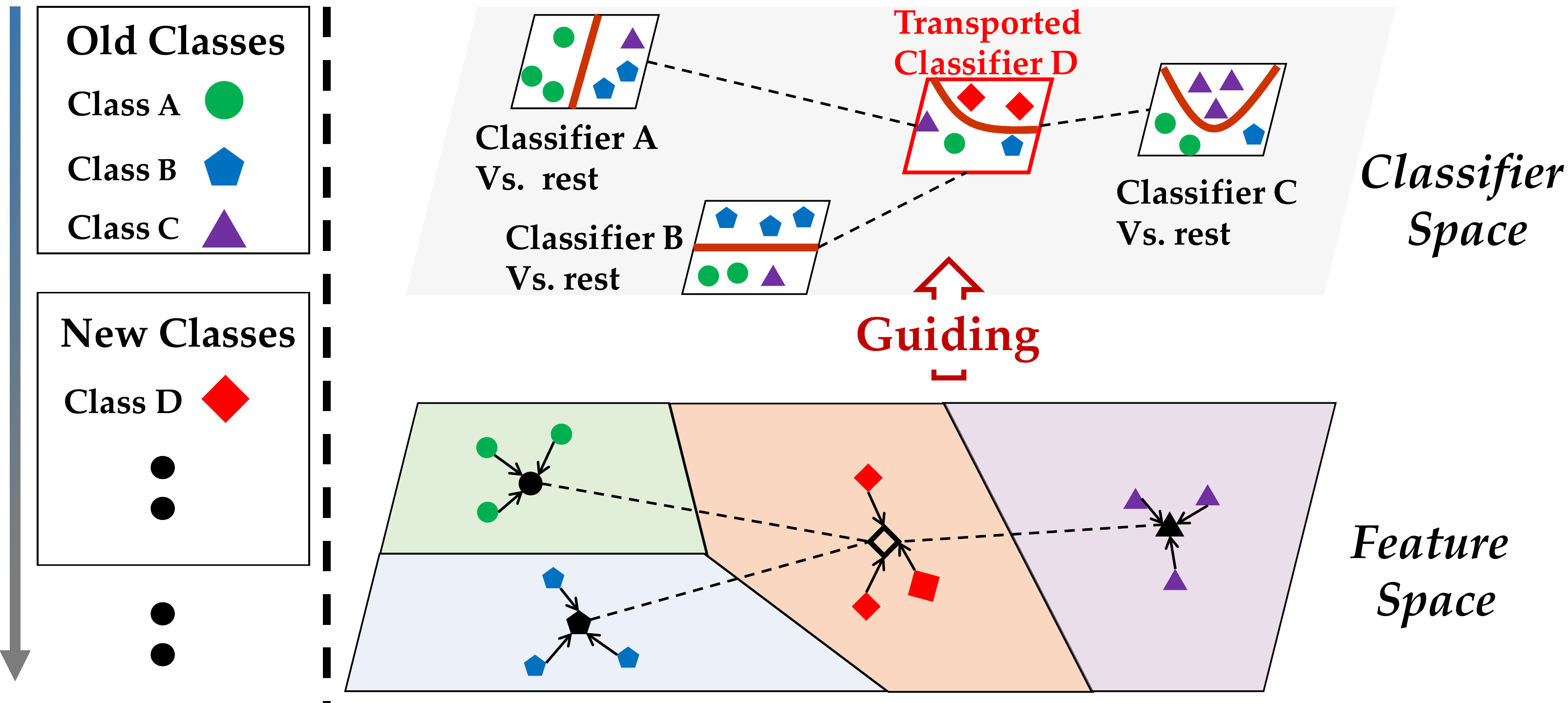}
	\end{center}
	
	\vspace{-3mm}
	\caption{\small  Left: setting of class-incremental learning. 
		We need to build a classifier for all the classes incrementally.
		Right: We can extract \emph{semantic relationship} in the feature space, and transfer classifiers of related classes to build that of new classes without training. Each classifier of old classes can be viewed as a one vs. rest classifier, and the class-wise relationship guides the synthesis of the new classifier.
	} \label{figure:intro}
	\vspace{-3mm}
\end{figure}

Correspondingly, we find that there is relevancy between old and new classes, \ie, \emph{semantic relationship}\footnote{Semantic relationship refers to the broader high-level information between concepts, \eg, word embedding in WordNet. Since we are unable to get such auxiliary information in the incremental data streams, we focus on the feature-wise relationship and treat it as the semantic relationship in this paper.}, 
benefited from which the linear classifier for old classes can be transported to one for new classes easily.
Figure~\ref{figure:intro} shows the schematic that the visual similarity among classes indicates the relationship among their corresponding linear classifiers with optimal transport~\cite{villani2008optimal,kantorovich1960mathematical}. We consider two different spaces, \ie, feature space and classifier space. We measure the relationship of classes in the feature space, and use the class-wise relationship to guide the classifier synthesis in the classifier space.
With the transported knowledge across different classes, the training process of old classes can facilitate new ones and vice versa.

Motivated by the fact that semantic relationship helps knowledge transfer, we propose CO-transport for class Incremental Learning (\mame), which learns to relate  across incremental tasks with the class-wise semantic relationship. 
In detail, the transport occurs in two stages. Facing the incoming of new classes, the model should quickly adjust and depict them. We propose \emph{prospective transport} to transfer old classifiers as initialization of new classes, which also helps  preserve the inner class relationship in new classes training. Additionally, to eliminate  the catastrophic forgetting, new classifiers should be transferred backward as old ones, acting as a distillation target. As a result, \emph{retrospective transport} is proposed to utilize the backward transferred knowledge, and prevent forgetting among old classes.
\emph{Prospective} and \emph{retrospective transport}  formulate the co-transport framework and transfer knowledge across multiple batches of incremental classes.
Vast experiments on benchmark and real-world multimedia datasets under various settings are conducted, which validate the effectiveness of \mame.

We start with a review of related work, followed by \name and experimental results. After that, we conclude the paper.

\section{Related Work}

Class-incremental learning (CIL)  is now a popular topic in the machine learning community~\cite{de2019continual,Zhang_2021_CVPR,Cheraghian_2021_CVPR,Liu_2021_CVPR,Wu_2021_CVPR,tao2020few,Zhu_2021_CVPRb,Zhu_2021_CVPR,Hu_2021_CVPR}. There are  two main ways for CIL: memory-based methods save old class exemplars to overcome forgetting, while non-memory based methods utilize regularization terms or consider dynamic model extension.

{\bf Non-memory based CIL:} Some methods expect network outputs do not drift sharply when training a new task, thus preserving the former knowledge. EWC~\cite{kirkpatrick2017overcoming} measures the importance of parameter to task by Fisher information matrix. EWC expects small changes in the important parameters, and regularizes them not shifting too much. SI~\cite{zenke2017continual} and MAS ~\cite{aljundi2018memory} follow EWC to regularize important parameters with different estimation. However, facing a long stream of incremental tasks, the importance matrix of different stages may conflict, making these algorithms perform poorly.
Another line of work changes the network structure to meet the requirements of new tasks. 
~\cite{yoon2018lifelong,xu2018reinforced} retrain and expand the  network for new tasks.
However, most methods need an oracle to guide which sub-network to be activated during the testing process. Additionally, the network expands with tasks emerging, the number of parameters becomes very large when facing long incremental sequences.

{\bf Memory-based CIL:} This line of work saves prototype/exemplar instances or prototype representations of former tasks in the memory. A rehearsal process will then be applied with these saved terms when learning a new task. iCaRL~\cite{rebuffi2017icarl} selects exemplars for replay and utilizes knowledge distillation of the former model to prevent forgetting. \cite{IscenZLS20} proposes to select features instead of images, and reduces the storage cost. 
Additionally, generative models can be viewed as another way to store exemplars. In~\cite{xiang2019incremental}, old class instances are synthesized to replay when training new tasks, and the incremental training process can be transferred into an offline training process. 
Recent works focus on compensating for the drift of the incremental model. \cite{wu2019large} utilizes the exemplars to build an extra validation set, with which an extra bias correction layer is trained. \cite{hou2019learning} proposes to utilize a cosine linear layer without bias to overcome forgetting. 
\cite{zhao2020maintaining,belouadah2019il2m} simply normalize the linear layer with weight clipping to maintain fairness between old and new classes. ~\cite{yu2020semantic} estimates the semantic drift of former class centers through new class instances. 
\name follows the memory-based line, and utilizes the class-wise semantic relationship across different incremental stages to reuse the old model, which is neglected by former approaches.

{\bf CIL for multimedia:} It is common to observe the emergence of new classes in real-world applications~\cite{zhou2016learnware,zhou2021learning,zhou2021detecting,ye2021contextualizing,wei2019multiple,ye2021learning,ning2021badge,yang21,lu2021tailoring,yang2015auxiliary}. As a result,
CIL has been found useful in vast multimedia fields, \eg, e-commerce product search~\cite{wang2020metasearch}, video action recognition\cite{yang2019}, natural language generation~\cite{mi2020continual}, multimedia retrieval~\cite{tian2020complementary}, and social media topic mining~\cite{xu2019open}.

{\bf Optimal transport (OT):} OT~\cite{villani2008optimal,peyre2019} is first formulated to study the optimal transportation and resource allocation problem~\cite{kantorovich1960mathematical,monge1781memoire}. With the ground cost, OT can find a coupling between two distributions, which can be viewed as the mapping between two sets~\cite{kolouri2017optimal,villani2008optimal}. 
The original OT computation involves the resolution of a linear program with a prohibitive cost, which is hard to implement. However, with the smoothness of entropy regularization term,
OT can be solved through Sinkhorn algorithm~\cite{cuturi2013sinkhorn,sinkhorn1967concerning}, which is much faster.
OT is now widely applied to machine learning and computer vision fields, \eg, model fusion~\cite{singh2020model}, 
domain adaptation~\cite{courty2016optimal}, object detection~\cite{ge2021ota}, model reuse~\cite{ye2020heterogeneous,ye2018rectify}, and generative models~\cite{arjovsky2017wasserstein,balaji2020robust}.

\section{From Old Classes to New Classes}
In this section, we first introduce the definition of CIL, followed by a typical memory-based approach.
After that, we discuss the insufficiency of the current model.
\subsection{Class Incremental Learning}
Class-incremental learning was proposed to learn a stream of data incrementally from different classes~\cite{rebuffi2017icarl}. Assume there are a sequence of $B$  training tasks $\left\{\D^{1}, \D^{2}, \cdots, \D^{B}\right\}$ without overlapping classes, where $\D^{b}=\left\{\left(\x_{i}^{b}, y_{i}^{b}\right)\right\}_{i=1}^{n_b}$ is the $b$-th incremental step with $n_b$ instances. Besides,  $\x_i^b \in \R^D$ is a training instance of class $y_i \in Y_b$, $Y_b$ is the label space of task $b$, where
$Y_b  \cap Y_{b^\prime} = \varnothing$ for $b\neq b^\prime$. 
During the training process of task $b$, we can only access data from $\D^b$.  
The aim of CIL at each step is not only to acquire the knowledge from the current task  $\D^b$, but also to preserve the knowledge from former tasks. 
After each task, the trained model is evaluated over all seen classes $\mathcal{Y}_b=Y_1 \cup \cdots Y_b$.

Each time a new task $\D^b$ arrives, the model should learn to classify the new classes among them. Assume the current model $f(\x)$ trained on $\D^{b-1}$  is composed of two parts: embedding  function $\phi(\cdot):\mathbb{R}^{D} \rightarrow \mathbb{R}^{d}$ and linear classifier\footnote{We omit the bias term, and use a cosine classifier. (See Sec.~\ref{sec:imp}).} $W_{old}\in\mathbb{R}^{d\times |\mathcal{Y}_{b-1}|}$, \ie, $f(\x)=W_{old}^{\top}\phi(\x)$. 
We denote the softmax operator as $\mathcal{S}(\cdot)$, and the predicted probability of class $k$ as $\mathcal{S}_k(W_{old}^{\top}\phi(\x))$.
The incremental model would first augment the linear classifier: $W_b=[W_{old}, W_{new}]$, where $W_{new}\in\mathbb{R}^{d\times |Y_b|}$ is \emph{randomly initialized} for new classes. Then the model is learned to predict over all the $|\mathcal{Y}_b|$ classes.

\subsection{CIL via Knowledge Distillation}

As we stated before, memory-based methods utilize a \emph{tiny} subset of old class exemplars to prevent forgetting, say $\mathcal{E}_{b-1}$, which is selected from $|\mathcal{Y}_{b-1}|$. A straightforward way to utilize these exemplars is to rehearsal and calculate cross-entropy:
\begin{align} \label{eq:crossentropy} 
	\mathcal{L}_{C E}(\mathbf{x}, y)=	 \sum_{k=1}^{|\mathcal{Y}_b|}-\mathbb{I}(y=k) \log \mathcal{S}_k(W_{b}^{\top}\phi(\x))
\end{align}
where $\mathbb{I}(\cdot)$ is the indicator function. Eq.~\ref{eq:crossentropy} optimizes the cross-entropy over all exemplars and novel instances, thus gains the knowledge and meanwhile resists forgetting.
However, $\mathcal{L}_{C E}$ is not enough to resist  forgetting, since the number of exemplars is much less than novel ones, \ie, $|\mathcal{E}_{b-1}| \ll |\D^b|$.
Hence, we need to align the prediction of old  and new model  through knowledge distillation~\cite{hinton2015distilling}:
\begin{align} \label{eq:kd}
	\mathcal{L}_{KD}(\mathbf{x}) =  \sum_{k=1}^{|\mathcal{Y}_{b-1}|}-
	\mathcal{S}_k(\bar{W}_{old}^{\top}\bar{\phi}(\x))
	\log \mathcal{S}_k(W_{b}^{\top}\phi(\x)) 
\end{align}
where $\bar{W}_{old}$ and $\bar{\phi}$ correspond to frozen classifier and embedding before learning $\D^b$.
The KD loss\footnote{We omit the temperature scalar $\tau$ for simplification. } maps the output of current model to the former model's output over all old classes. 
The aligned probabilities make the current model have  the same  discrimination as the old one, thus restrict former knowledge from forgetting. 
The overall loss combines $\mathcal{L}_{CE}$ and $\mathcal{L}_{KD}$:
\begin{align} \label{eq:icarl}
	\mathcal{L}(\x, y)=(1-\lambda) \mathcal{L}_{CE}(\x, y)+\lambda \mathcal{L}_{KD}(\x)
\end{align}
where $\lambda$ is a trade-off parameter to balance the importance of new and old classes, which is set to $\frac{|\Y_{b-1}|}{|\Y_b|}$ ~\cite{wu2019large}.

\subsection{Ignorance of Semantic Relationship }
Eq.~\ref{eq:icarl} depicts a way to utilize exemplars for CIL through rehearsal and knowledge distillation. However, there is some side information neglected in the learning process. Firstly, facing the new classes, the linear classifier $W_b=[W_{old}, W_{new}]$ is a simple augmentation  with randomly initialized $W_{new}$, which may negatively affect the current model. 
Correspondingly, there exists a mapping between old and new classes, \ie, \emph{semantic relationship}. Since the old and new models are related in classes, we should not neglect the current model, but rather utilize the class-wise similarity to assist CIL. 
With the increasing of old classes, the probability of new classes related to old ones also increases, making the transfer easier.
Furthermore, current methods transfer knowledge in the single direction of knowledge flow, \ie, from the old model to the new model. It is promising to use semantic information as further guidance
and transport knowledge retrospectively, thus preserving old knowledge.

\section{Co-Transport via Optimal Transport}
Motivated by the informative \emph{semantic relationship}, we seek to relate old and new classes via model reuse. Furthermore, the embedding module $\phi(\cdot)$ is generalizable, capturing input's common features and clustering them with learned metrics, thus unrelated to classes. In contrast, the linear layers $W$ are directly related to the classes. As a result, we should transfer and reuse the linear layers based on the current embedding. 
Suppose we already extract the semantic relationship between  classes; we propose a semantic mapping $\T$ \emph{which transfers a linear classifier from origin to the goal classes}.  $\T$ takes the origin classifier as input and produces a well-suited classifier for the goal classes. With semantic mapping, we can transport the old classifier for the new classes when they arrive. The generated classifier will not suffer the negative influence of random initialization, but rather act as a promising starting point for new classes. 
Symmetrically, we can transport the new classifiers to the old classes when learning new tasks, and encourage knowledge preserving with the transported one. As a result, the knowledge in the model is transported in two directions, \ie, prospectively and retrospectively, and the framework is thus named as co-transport.

Suppose we already know the expression of $\T$, we first introduce the co-transport framework, and at last obtain the transformation $\T$ via optimal transport.

\subsection{Prospective transport (PT)}
Facing a new task, the model should adapt quickly towards new classes. PT solves the model adaptation with semantic mapping $\T$ across old and new classes.
For example, suppose we have the well-trained weights to predict `cat' in the old classes. In that case, we can reuse almost exactly the same classifier to determine class `tiger.' As a result, we build the new classifier $W_{new}$ reusing the old ones $W_{old}$, directed by the semantic mapping $\T$: $\bar{W}_{new}= \T(W_{old})$. PT-guided classifier helps to learn new classes in two aspects:

\noindent {\bf Fast initialization:} Comparing to the randomly generated new classifier, the transferred $\bar{W}_{new}$ well utilizes the former one, and meanwhile preserves the semantic relationship across classes. The calibration between old and new classes is maintained because semantic mapping captures the class-wise relationship. As a result, the transported classifier can tell  the new classes apart even not trained with them. $\bar{W}_{new}$ acts as a good initialization for new classes, without the negative influence of random initialization.

\noindent {\bf PT loss:} Since $\bar{W}_{new}$ preserves the class-wise relationship, it is helpful to adjust the predicted probability towards the transferred model at the very beginning.
We exert an extra restriction to the model via knowledge distillation to achieve this goal:
\begin{align}
	\label{eq:pt}
	\mathcal{L}_{PT}(\mathbf{x}) = \sum_{k=1}^{|\mathcal{Y}_{b}|}-
	\mathcal{S}_k\left({\left[\bar{W}_{old},\bar{W}_{new}\right]^{\top}\bar{\phi}(\x)}\right)
	\log \mathcal{S}_k\left({W_{b}^{\top}\phi(\x)}\right) 
\end{align}
where $\bar{W}_{new}$ is the initialized value. Eq.~\ref{eq:pt} forces the current updating model to predict like the transformed one, and helps the updating model 
to fit for new classes in a supervised manner.

\noindent {\bf Effect of PT:} Taking class-wise semantic relationship into consideration, PT reuses the old  classifier to build the new ones, and avoids the negative effect of random initialization.
PT aims for a better initialization of the parameters. Like~\cite{arnold2021maml}, a good initialization helps learn a better embedding, which maintains the former knowledge and resists catastrophic forgetting.
When there exists semantic relation among old and new classes, the transported classifier can incorporate and calibrate the new classes into the current classifier.
{Even if old and new classes are irrelevant, such weak initialization will not perform worse than the random initialized one.} The fast initialization can predict correctly even without trained on new classes (Sec.~\ref{sec:vis}). In the implementation, we utilize PT loss for class-wise calibration at the beginning, \eg, first five epochs.

\begin{figure}[t]
	\begin{center}
		{\includegraphics[width=1\columnwidth]{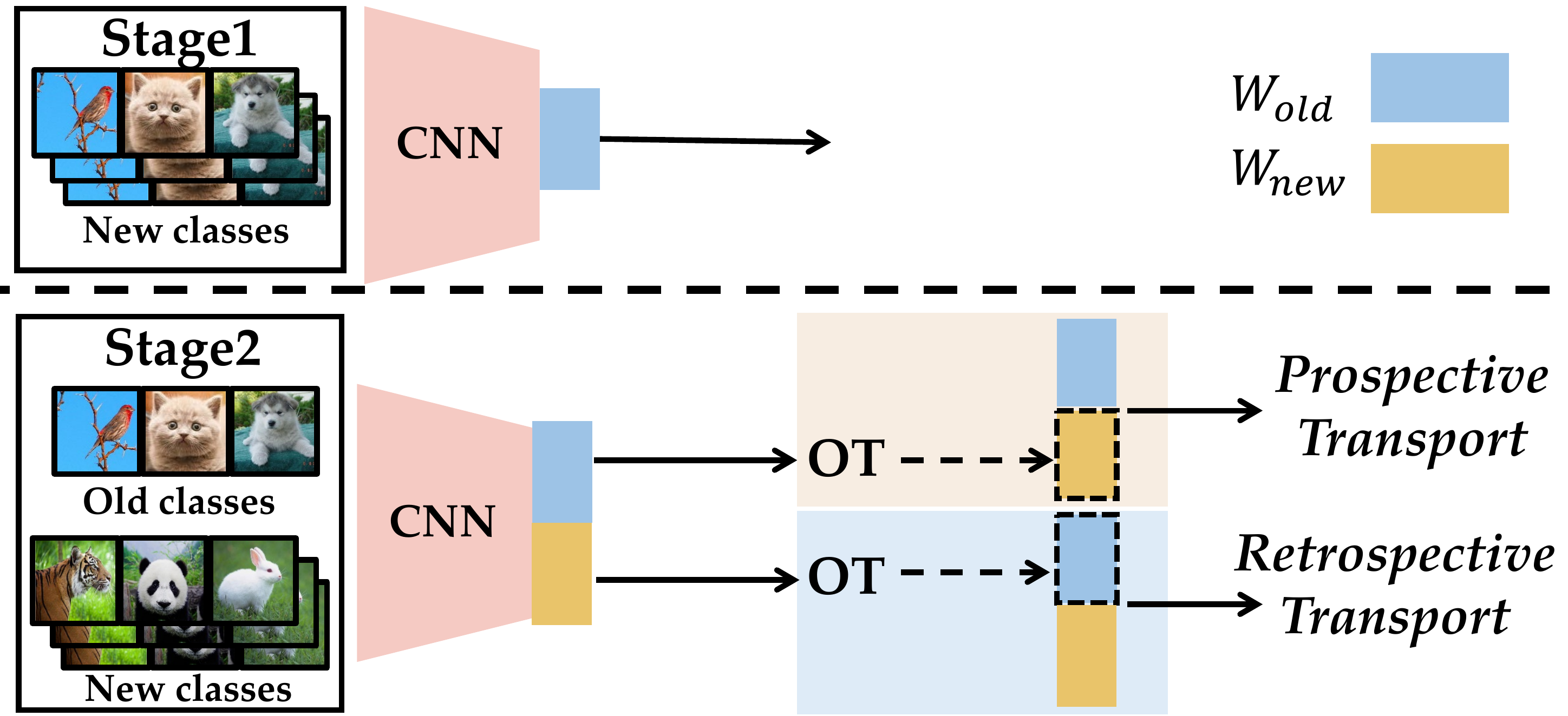}\label{figure:2a}}
	\end{center}
	\vspace{-3mm}
	\caption{\small Illustration of  \name framework. With OT guided \emph{semantic mapping}, classifiers can be transported across different stages. The yellow color indicates prospective transport, which transfers the old classifier as new ones as initialization. The blue color indicates retrospective transport, which transfers the new classifier as old ones to overcome forgetting. Dotted boxes denote transported classifiers, and we use their output as a distillation term to enable knowledge transportation.
	} \label{figure:method}
	\vspace{-3mm}
\end{figure} 

\subsection{Retrospective transport (RT)}
In addition to prospective transport, which reuses old classifiers to build new ones, we propose that the semantic relationship contained in the updating model can be transported  retrospectively. New classifiers can also be transported as old ones with the new-to-old mapping, \ie, $\hat{W}_{old}= \T(W_{new})$.  The semantic mapping now acts as an element to build a regularization of former class knowledge.

\noindent {\bf RT loss:} Similar to PT loss, we construct RT loss with the transformed classifier to refrain from forgetting:
\begin{align} \label{rt}
	\mathcal{L}_{RT}(\mathbf{x}) = \sum_{k=1}^{|\mathcal{Y}_{b-1}|}-
	\mathcal{S}_k\left({\bar{W}_{old}^{\top}\bar{\phi}(\x)}\right)
	\log \mathcal{S}_k\left({\hat{W}_{old}^{\top}\phi(\x)}\right)  
\end{align}
which forces the transformed model to predict similar results as  the old model and maintain the discrimination ability. RT loss builds a mapping to preserve former knowledge over the old classes. Comparing Eq.~\ref{rt} to Eq.~\ref{eq:kd}, the difference lies in the distillation student changes into transformed model $\hat{W}_{old}^{\top}\phi(\x)$. The transformation mapping $\T$ guides the preserving process with semantic relationship.

\noindent {\bf Effect of retrospective transport:} RT utilizes the semantic mapping to transform the new classifier into the old classifier, and restricts the class-wise relationship through knowledge distillation. A more robust regularization is thus exerted over the model and prevents forgetting. Note that current methods only transfer knowledge in a \emph{single} direction, while RT \emph{enables two classifiers to co-supervise each other and bi-directionally transfer}.
In the implementation, we gradually increase the weight of RT loss with epochs increasing. 

\noindent {\bf Summary of Co-Transport:} Figure~\ref{figure:method} illustrates the \name framework. Facing the incoming new classes, we extract the class-wise relationship and transport knowledge  $W_{old}\rightarrow \bar{W}_{new}$, which facilitates model adaptation with old classes. During the learning process of new classes, we extract the semantic relationship and transport knowledge  ${W}_{new}\rightarrow \hat{W}_{old}$, which helps overcome forgetting. These two transportation cooperate with each other with a knowledge flow \emph{in two directions}, thus  facilitating class-incremental learning.

{\begin{algorithm}[t]
		\caption{ Co-transport for incremental learning }
		\label{alg:coil}
		\raggedright
		{\bf Input}: New task instances: $\D^{b}$; former exemplars: $\mathcal{E}_{b-1}$;
		
		{\bf Output}: Updated model 
		\begin{algorithmic}[1]
			\State $W_{new}=\text{OT}(\mathcal{E}_{b-1},\mathcal{D}^{b},W_{old})$; \Comment{Fast initialization}
			\State $\mathcal{D}_{train}=\D^{b} \cup\mathcal{E}_{b-1}$; \Comment{Rehearsal}
			\Repeat
			\State Get a batch of instances;
			\State $\mathcal{L}_{PT}$$\leftarrow$Eq.~\ref{eq:pt}; \Comment{Prospective transport loss}
			\State $\hat{W}_{old}=\text{OT}(\mathcal{D}^{b},\mathcal{E}_{b-1},W_{new})$;
			\State $\mathcal{L}_{RT}$$\leftarrow$Eq.~\ref{rt}; \Comment{Retrospective transport loss}
			\State $\mathcal{L}_\text{\name}$ $\leftarrow$Eq.~\ref{eq:losscoil};\Comment{Total loss}
			\State SGD over $W_b$ and $\phi(\cdot)$;\Comment{Update model}
			\Until reaches predefined epochs
		\end{algorithmic} 
\end{algorithm}}

\subsection{Semantic Mapping via Optimal Transport}
For now, we have proposed the co-transport framework. The biggest problem left is: how to transform one classifier into another, \ie, how to get the semantic mapping $\T$?
$\T$ should capture the correlation across class sets, and capable of transforming an {original} classifier $W_{o}$ as a goal one $W_{g}$. 
A linear layer's coefficient reveals the positive/negative relationship between a feature and a class, and the prediction $W^\top \phi(\x)$ is the vector of class-specific weighted sum over all features. 
Therefore, we can \emph{reweight} the prediction of the transformed one with a linear mapping $T \in \R^{\alpha \times\beta}$. $T$ encodes the class-wise semantic correlation between a size-$\alpha$ class set of the original  task and the size-$\beta$ class set of the goal task. The more two classes are related, the larger the corresponding value in $T$. Since related classes rely on similar features to determine the label, we can reuse similar classes' weights in the origin task to get a good prediction in the goal task. For example, important features to predict `cat' can also help predict `tiger' in the goal task, and vice versa. With $T$ we can transport the class-wise relationship across different class sets. We then introduce how to calculate the mapping $T$.

\noindent {\bf Transportation mapping:} Denote $\boldsymbol{\mu}_{1} \in \Delta_{\alpha}$, $\boldsymbol{\mu}_{2} \in \Delta_{\beta}$, where
$\Delta_{d}=\left\{\boldsymbol{\mu}: \boldsymbol{\mu} \in \mathbb{R}_{+}^{d}, \boldsymbol{\mu}^{\top} \mathbf{1}=1\right\}$ is $d$-dimensional simplex. 
$\boldsymbol{\mu}_1$ and $\boldsymbol{\mu}_2$ are normalized marginal probability representing the importance of each class, which is set to uniform distribution without informative prior. 
A cost matrix $C\in \R_{+}^{\alpha \times \beta}$ is introduced to depict the class variation and guide the transition, whose elements point out the cost we should pay when link one class of the origin task to the goal counterpart. Thus we can rethink $T$ as a coupling of two distributions, which relates classes between tasks with lowest transportation cost, which can be optimized via minimizing:
\begin{align} \label{eq:ot}
	\min _{T}\langle T, C\rangle \quad \text { s.t. } T \mathbf{1}=\boldsymbol{\mu}_{1}, T^{\top} \mathbf{1}=\boldsymbol{\mu}_{2}, T \geq 0 
\end{align}
Eq.~\ref{eq:ot} is the Kantorovitch formulation of OT, where T shows how to align one set with another. The probability mass of a class will be moved to similar classes with small costs. Considering the arrival of new class, with uniform class marginals $\boldsymbol{\mu}_{1} \in \Delta_{|\Y_{b-1}|}$, $\boldsymbol{\mu}_{2} \in \Delta_{|Y_b|}$. Eq.~\ref{eq:ot} will output a permutation map $T$ with the right alignment between two class sets.
Applying this alignment of classes over models, we can transform a well-trained classifier from the former task to the current one. Eq.~\ref{eq:ot} is a middle function, which is not an optimization term.

\noindent {\bf Transportation cost:} In Eq.~\ref{eq:ot}, the given $C$ characterizes the relationship between two different class sets from origin and goal tasks. Instead of handcrafting the values of $C$, we propose a general approach to encode the invariant relationship across classes. We first encode all the classes in the same form.

{\begin{algorithm}[t]
		\caption{ OT for classifier transportation }	\label{alg:ot}
		\raggedright
		{\bf Input}: Origin exemplars: $\mathcal{E}_{o}$; 
		
		Goal exemplars: $\mathcal{E}_g$; 
		
		Origin classifier: $W_o$;
		
		{\bf Output}: Transported goal classifier ${W}_g$;
		
		\begin{algorithmic}[1]
			\Function{OT}{$\mathcal{E}_{o},\mathcal{E}_g,W_o$}
			\State Class center of embeddings $\leftarrow$ Eq.~\ref{eq:center};
			\State Transportation cost $C$ $\leftarrow$ Eq.~\ref{eq:cost};
			\State Solve OT problem in Eq.\ref{eq:ot}; get transport mapping $T$;
			\State	\Return ${W}_g=TW_{o}$
			\EndFunction
			
		\end{algorithmic}
		
\end{algorithm}}

In particular, we have the new classes from the new task and old ones from the exemplar set $\mathcal{E}_{b-1}$. 
The  class center of each class with the current embedding can be derived:
\begin{align} \label{eq:center}
	v_{n}= \frac{\sum_{i=1}^{|\mathcal{E}_{b}|} \mathbb{I}(y_i=n)\phi(\x_i) }{\sum_{i=1}^{|\mathcal{E}_{b}|} \mathbb{I}(y_i=n) } 
\end{align}
For simplification, we extract from all seen classes as $\mathcal{E}_{b}$. 
If two classes are related, their corresponding class-wise representations will also be close to each other. We use the pairwise Euclidean distance to measure the cost $C$ between classes:
\begin{align} \label{eq:cost}
	C_{n, m}=\left\|v_{n}-v_{m}\right\|_{2}^{2} 
\end{align}
the larger the distance, the more dissimilar between two classes. As a result, there takes more difficulty to reuse the particular coefficient of the previous well-trained model. Substituting Eq.~\ref{eq:cost} to Eq.~\ref{eq:ot}, the learned transportation plan $T \in \R_+^{\alpha \times \beta}$ directs how to transfer the classes of the origin task to the goal tasks domains with the lowest cost. We solve  OT with Sinkhorn algorithm~\cite{cuturi2013sinkhorn}.

\subsection{Guideline for Implementation} \label{sec:imp}
We show the guideline of \name in Alg.~\ref{alg:coil}. 
The overall loss  is a combination of Eq.~\ref{eq:icarl} with co-transport loss:
\begin{align} \label{eq:losscoil} 
	\mathcal{L}_\text{\name}(\x, y)= \sum_{\x \in \D^b \cup \mathcal{E}_{b-1}} \mathcal{L}(\x, y)+	\mathcal{L}_{PT}(\x) + \gamma	\mathcal{L}_{RT}(\x)
\end{align}
where $\gamma$ acts as a cumulative learning adjuster:
$\gamma=\left({t}/{T_{max}}\right)^2$, $t$ and $T_{max}$ stand for the current epoch index and total  epochs. $\mathcal{L}_{PT}$ only works at the first five epochs for fast model initialization.
Note that OT is a middle function, which is not an optimization term, and PT loss and RT loss are optimized simultaneously.
We use \emph{cosine classifier}, \ie, the features and weights of linear layer are both normalized before multiplied: $f(\x)=(\frac{W}{\|W\|_2})^\top(\frac{\phi(\x)}{\|\phi(\x)\|_2})$. As a result, we do not need to care about  the calibration  between old and new classes. 
See supplementary for more details.

\begin{figure*}[t]
	\begin{center}
		\subfigure[ CIFAR-100, 20 tasks]
		{\includegraphics[width=.5\columnwidth]{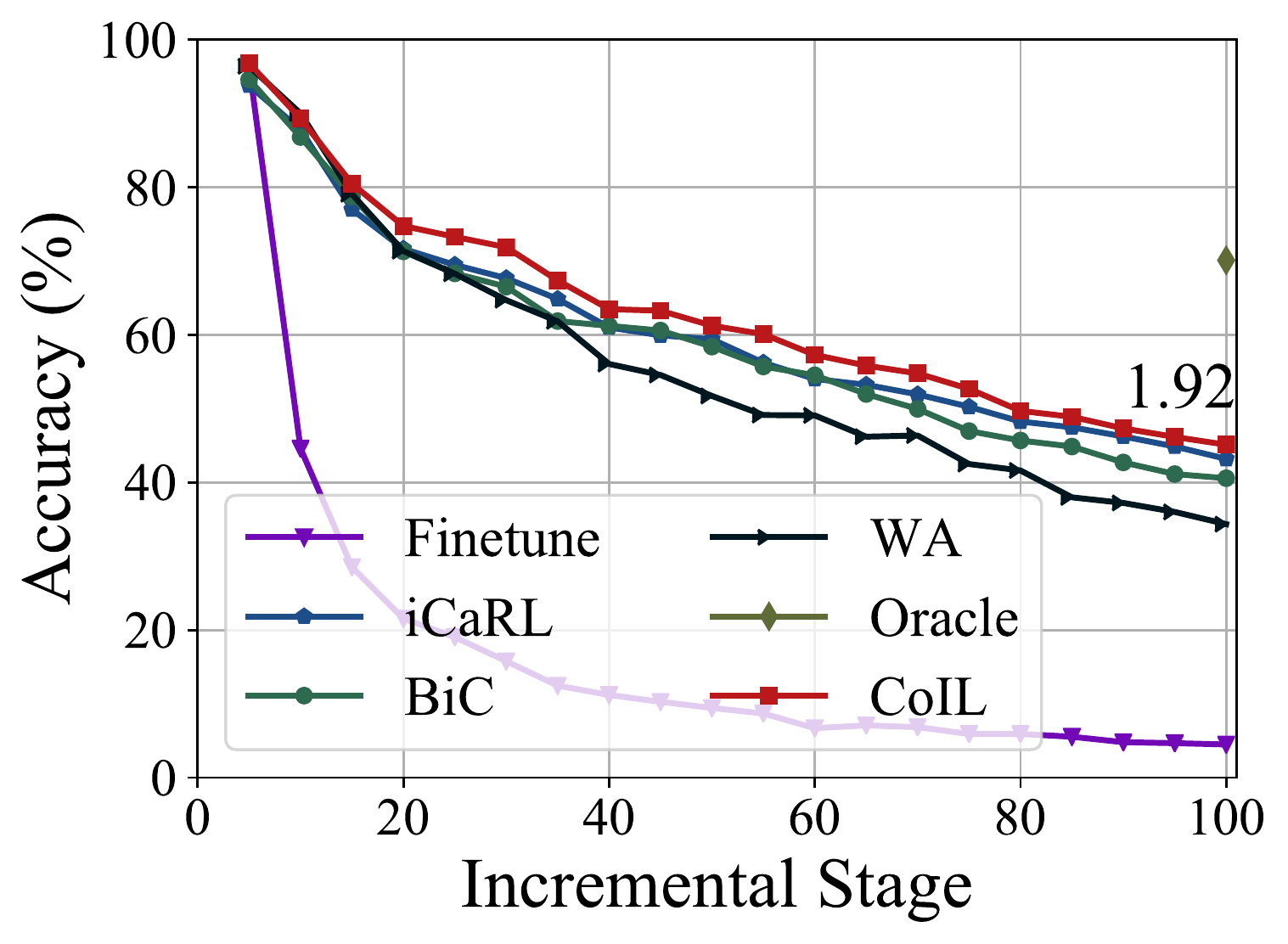}}
		\subfigure[ CIFAR-100, 10 tasks]
		{\includegraphics[width=.5\columnwidth]{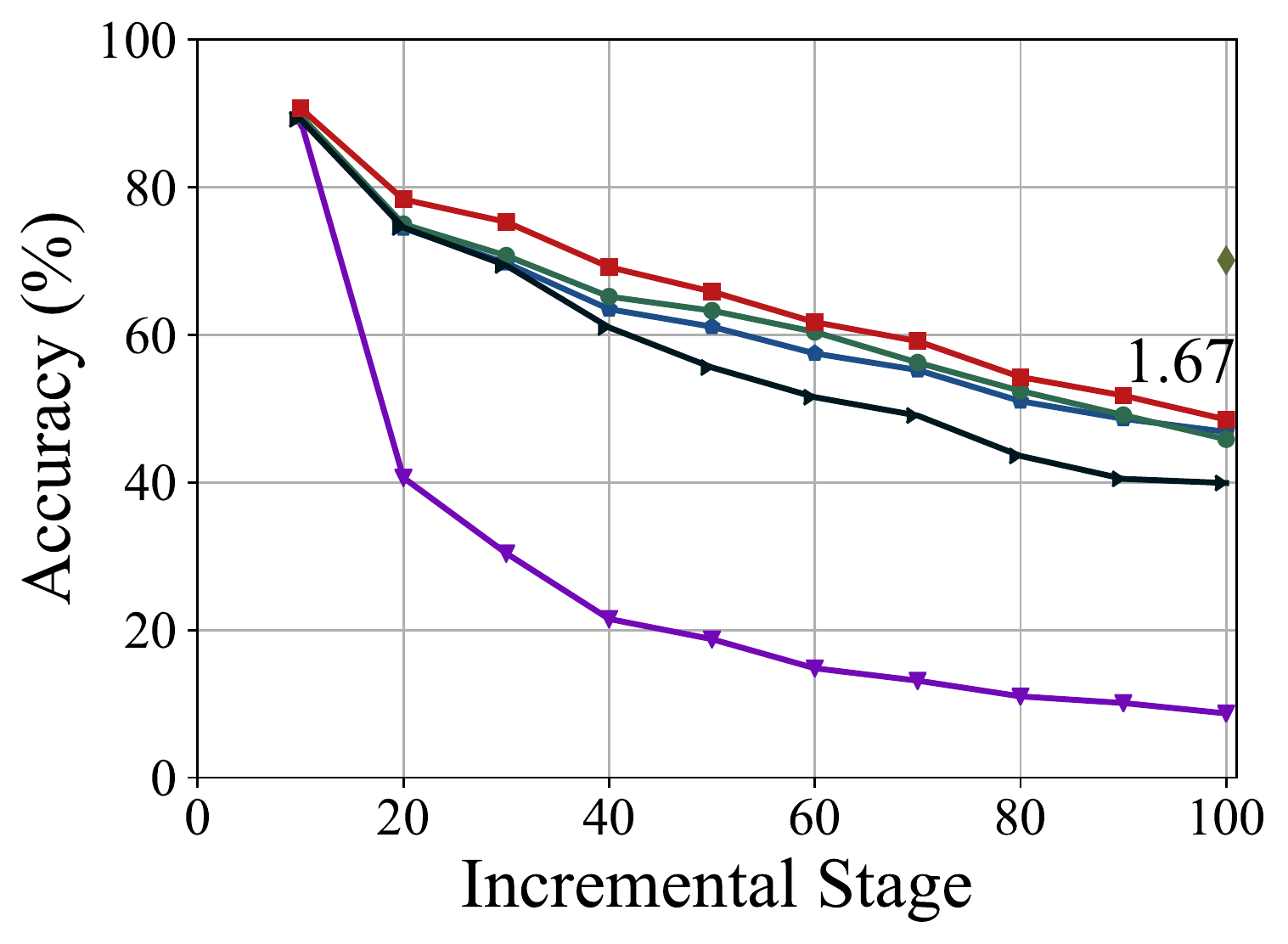}}
		\subfigure[ CIFAR-100, 5 tasks]
		{\includegraphics[width=.5\columnwidth]{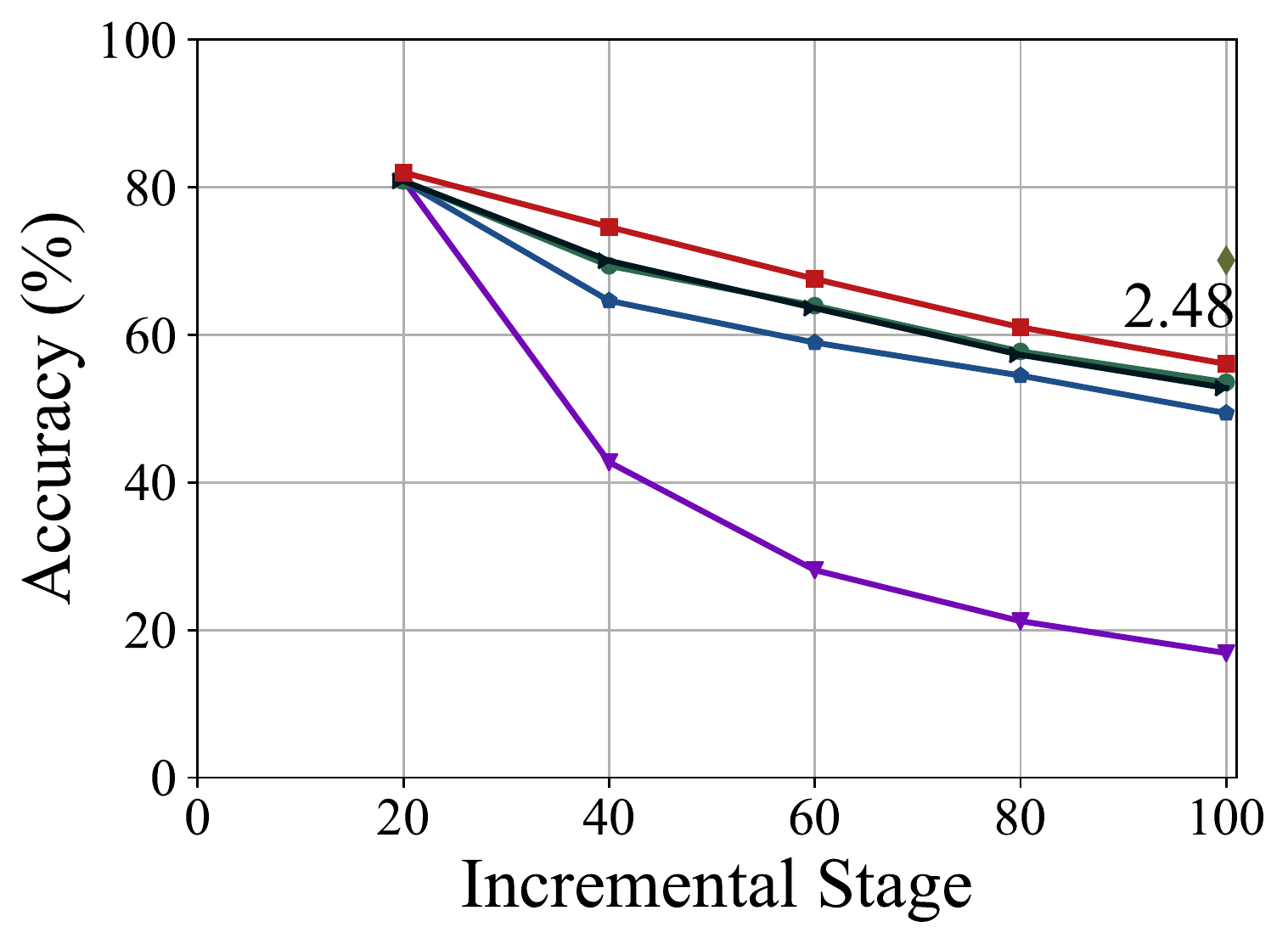}}
		\subfigure[ CIFAR-100, 2 tasks]
		{\includegraphics[width=.5\columnwidth]{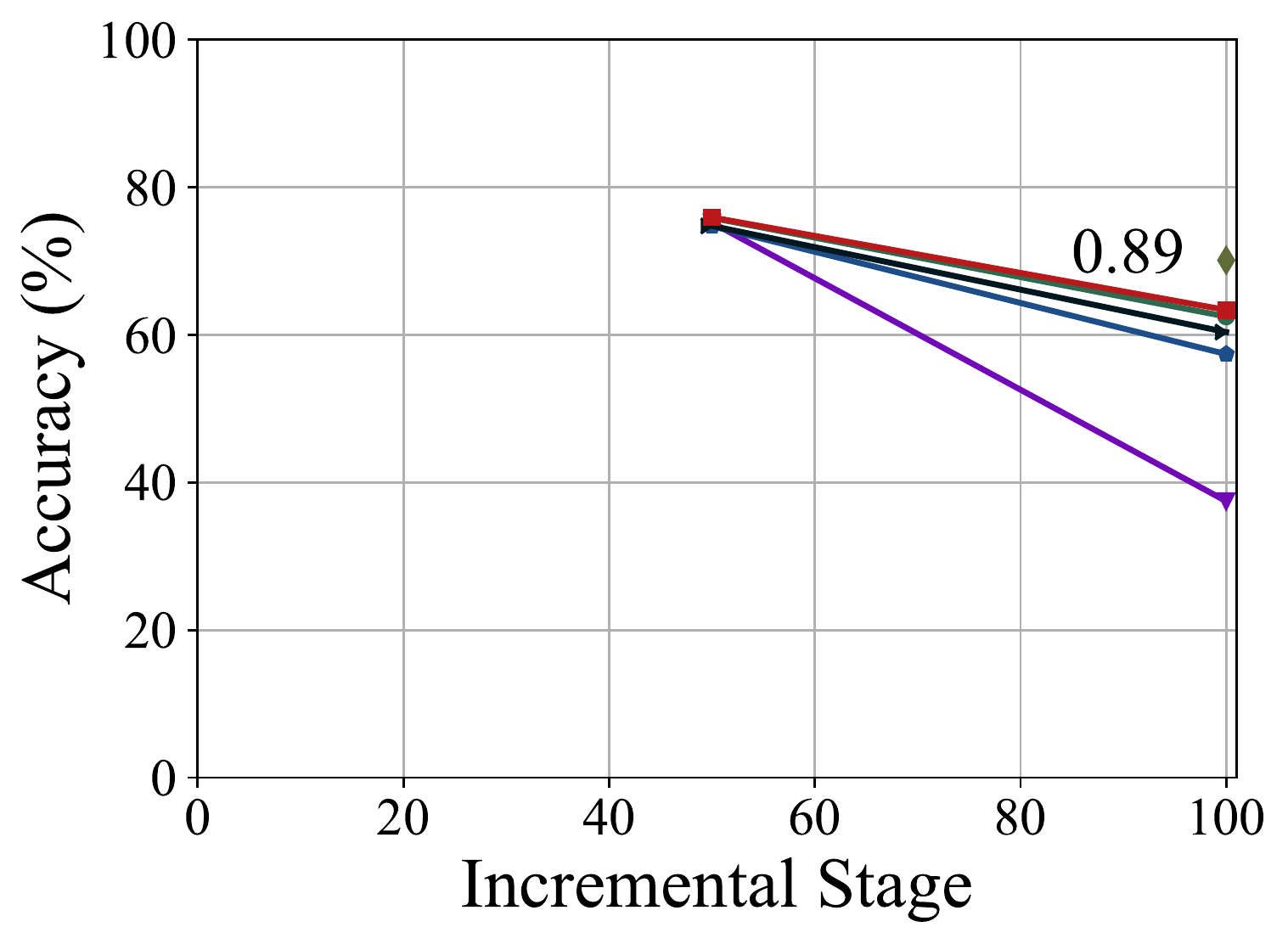}}
		\\ 
		\vspace{-3mm}
		\subfigure[ ImageNet-100, 10 tasks]
		{\includegraphics[width=.5\columnwidth]{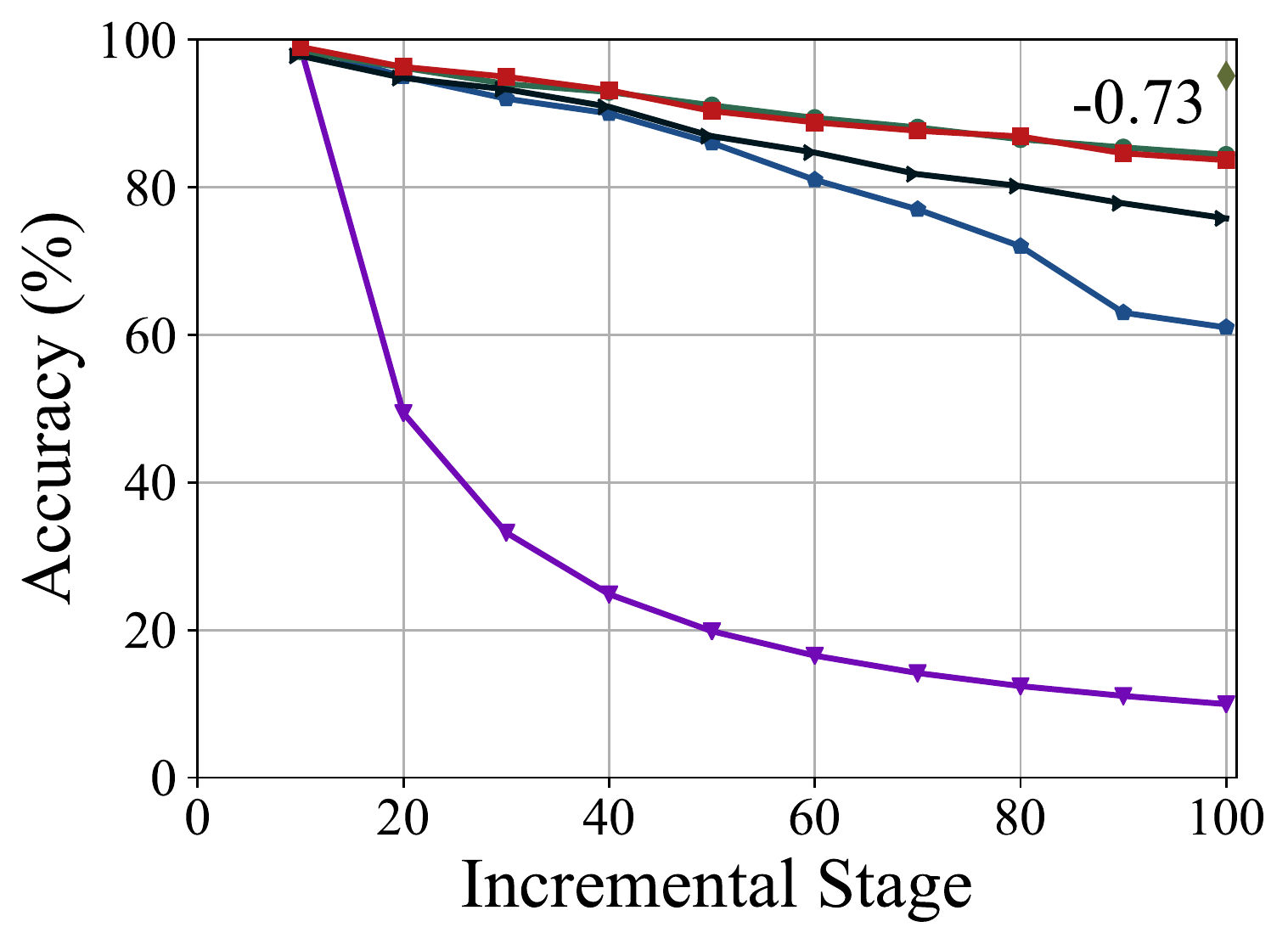}}
		\subfigure[ ImageNet-1000, 10 tasks]
		{\includegraphics[width=.5\columnwidth]{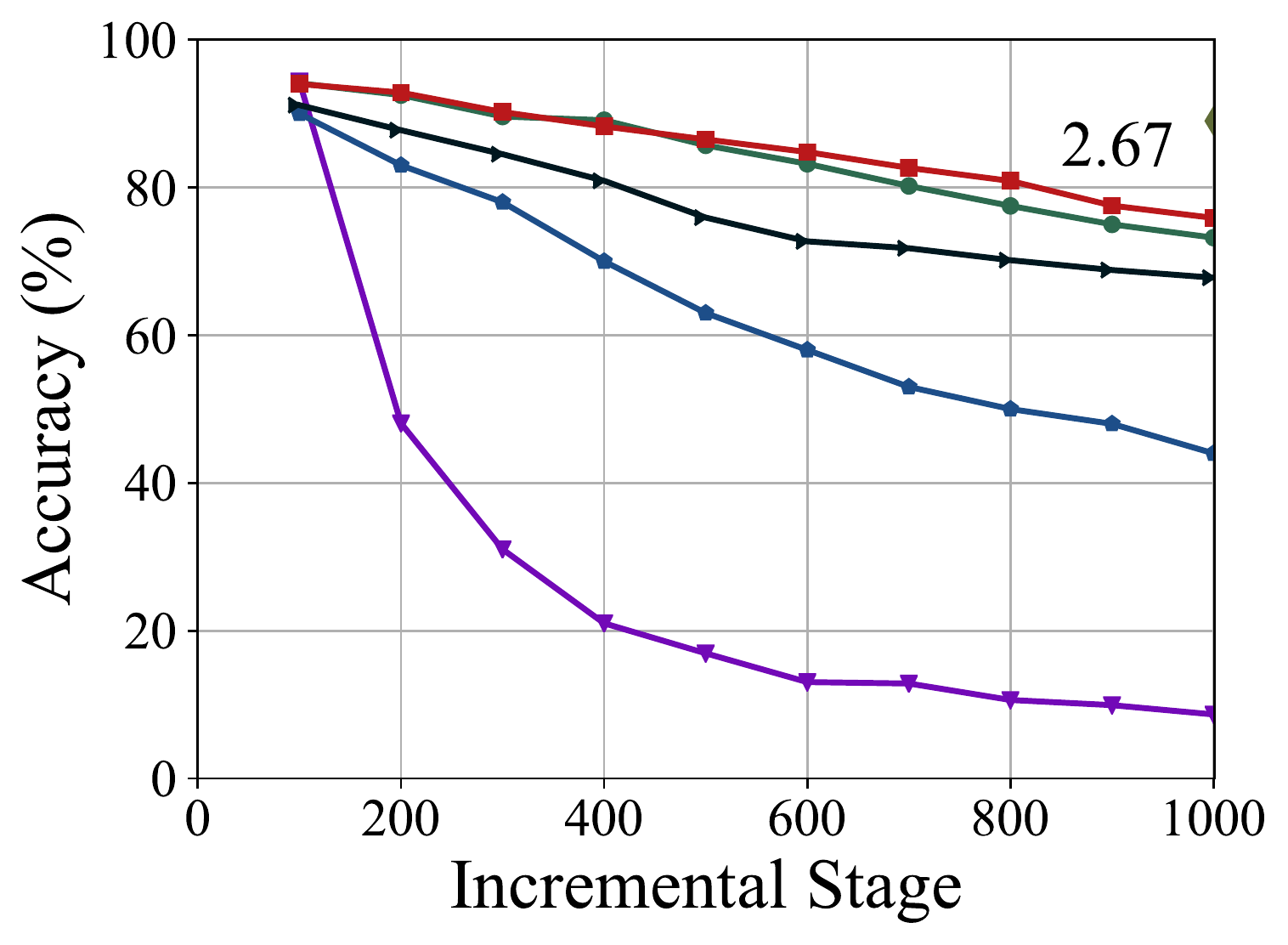}}
		\subfigure[ CUB-100, 10 tasks]
		{\includegraphics[width=.5\columnwidth]{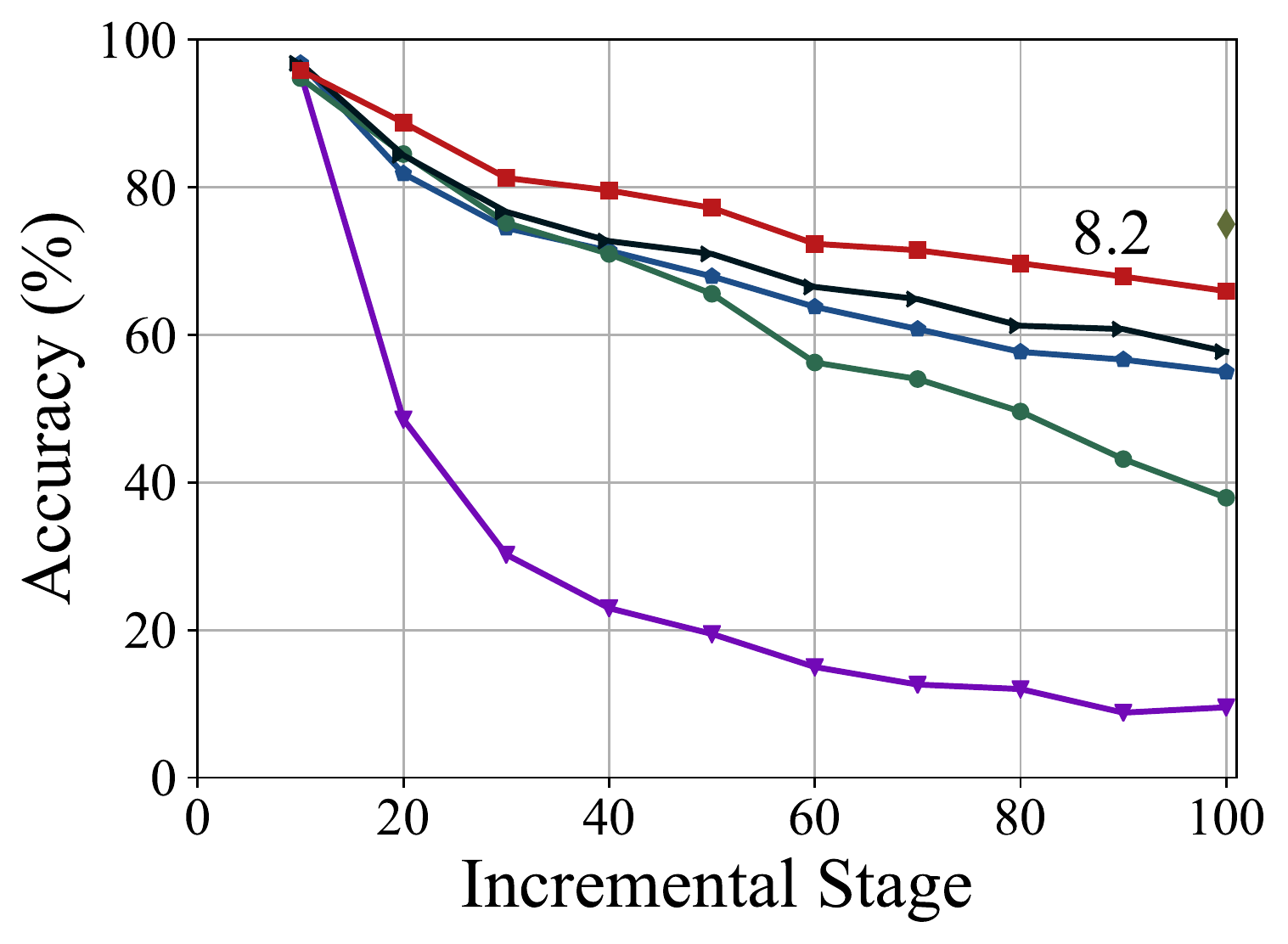}}
		\subfigure[ CUB-200, 10 tasks]
		{\includegraphics[width=.5\columnwidth]{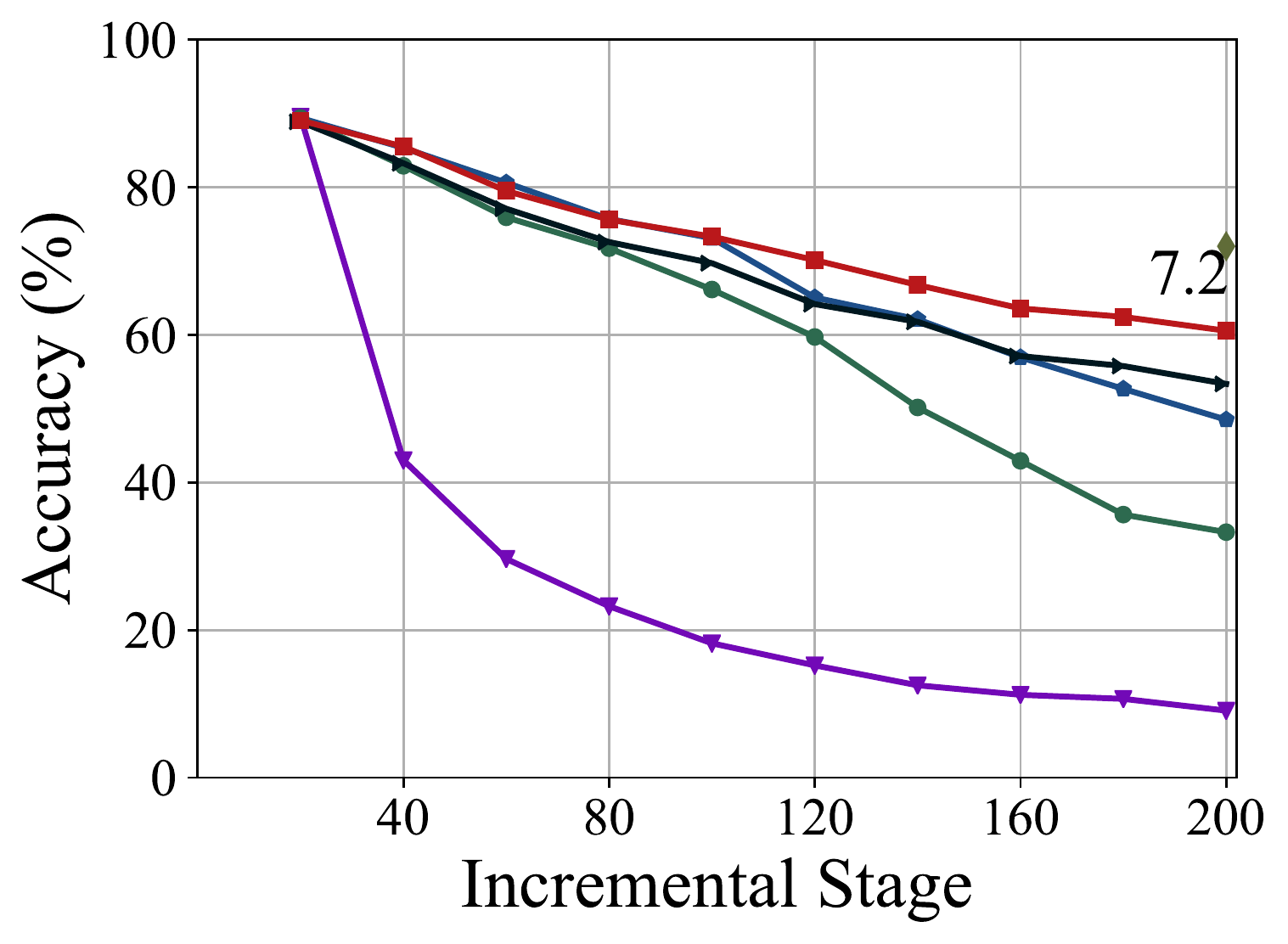}}
		\vspace{-3mm}
	\end{center}
	\caption{\small Performance on benchmark datasets, \ie, CIFAR-100, ImageNet-100/1000 and CUB-100/200. The legends are shown in (a), and the number at the end indicates how much our method outperforms the second one on the last incremental stage. The averaged accuracy comparison over all incremental stages is reported in Table~\ref{table:inc}.  }
	\vspace{-3mm}
	\label{figure:sota}	
\end{figure*}

\begin{figure*}[t]
	\begin{center}
		\subfigure[CIFAR-100, base 50 classes, 10 tasks]
		{\includegraphics[width=.65\columnwidth]{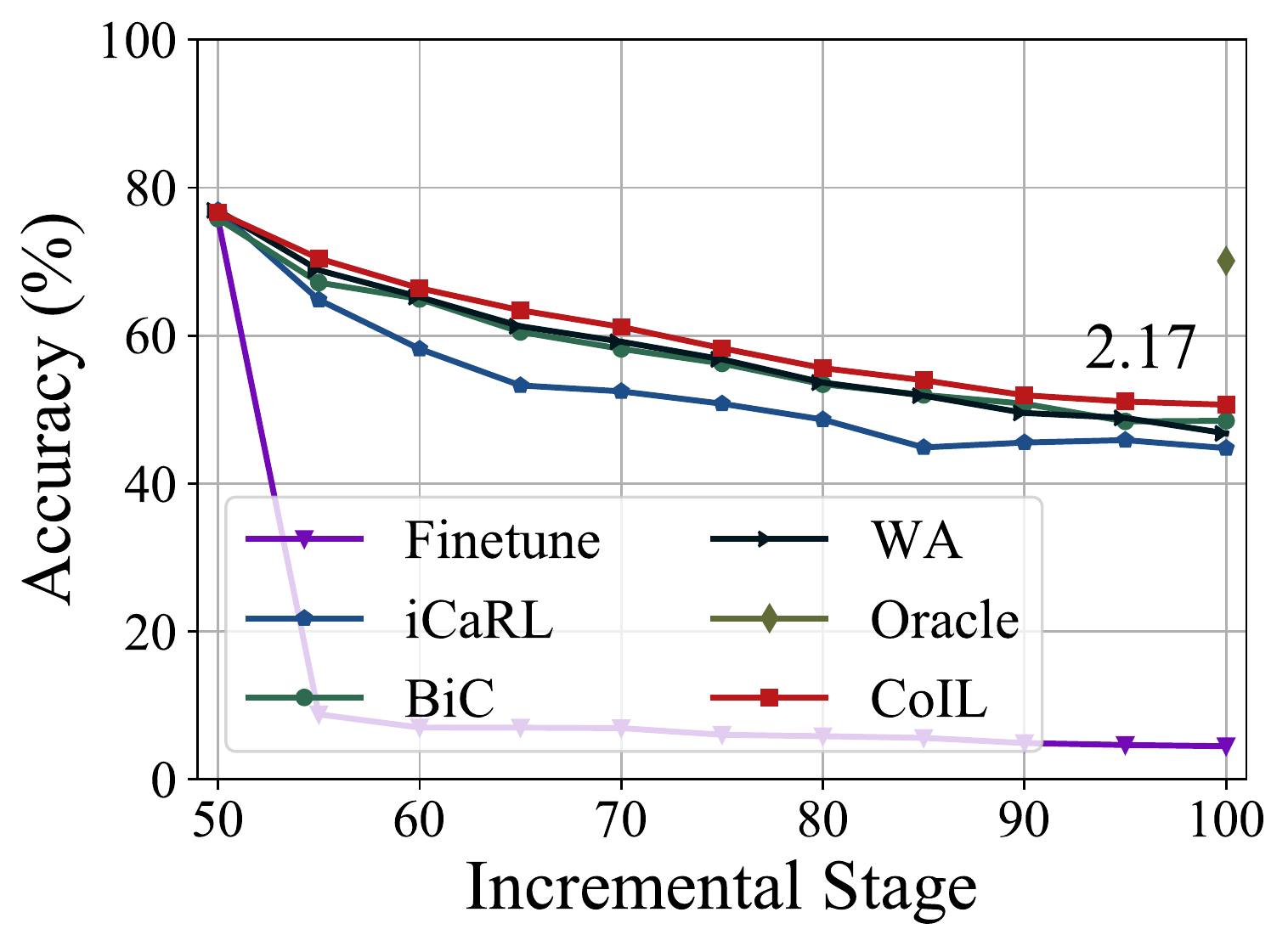}}
		\subfigure[ CIFAR-100, base 50 classes, 5 tasks]
		{\includegraphics[width=.65\columnwidth]{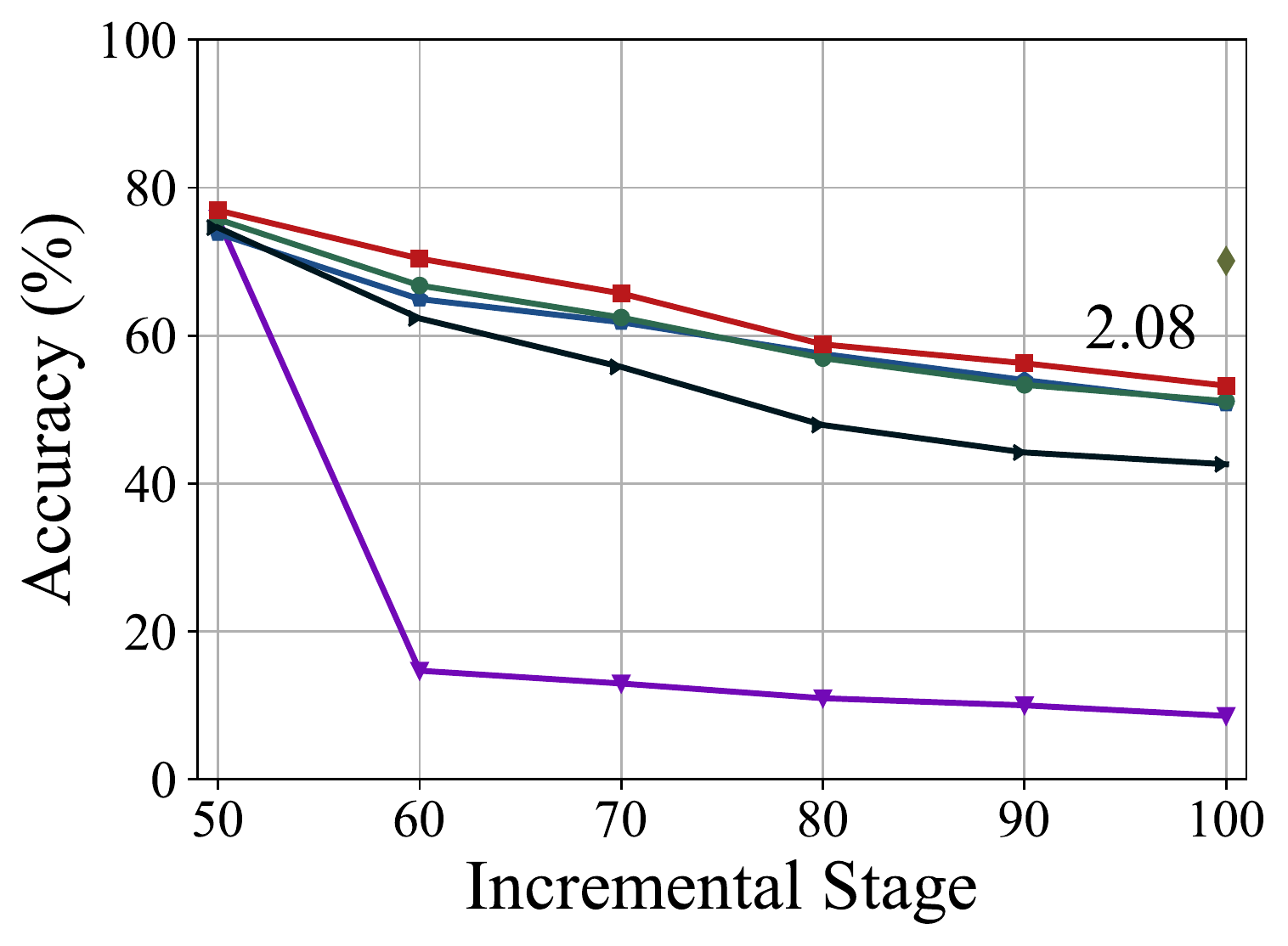}}
		\subfigure[ CIFAR-100, base 50 classes, 2 tasks]
		{\includegraphics[width=.65\columnwidth]{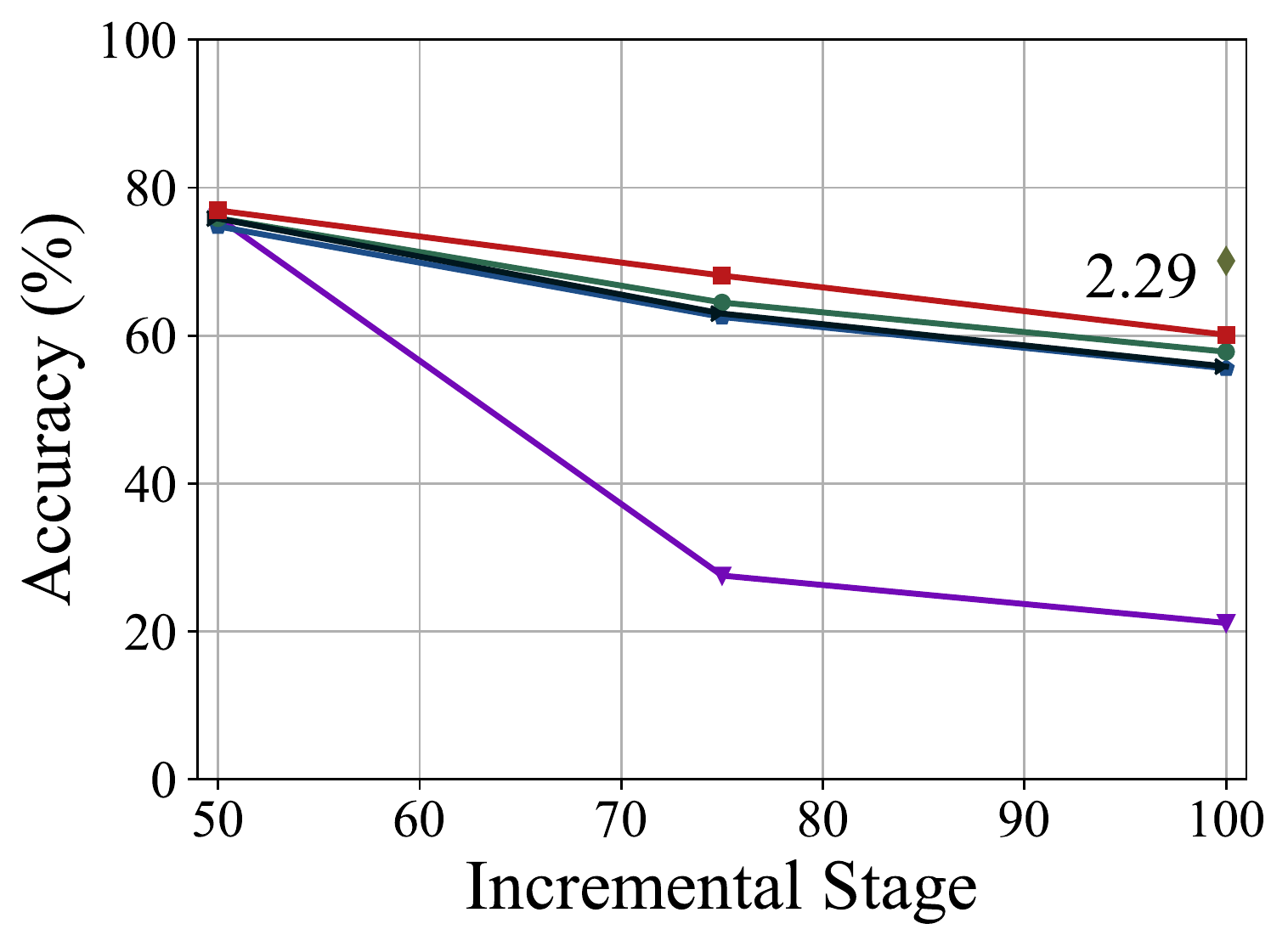}}
	\end{center}
	\vspace{-3mm}
	\caption{ \small Performance over CIFAR-100, half of the total classes are trained in the first task. The legends are shown in (a), and the number at the end indicates how much our method outperforms the second one on the last incremental stage.   } \label{figure:sotacifar}
	\vspace{-3mm}	
\end{figure*}

\section{Experiment}
In this section, we compare \name on benchmark and real-world incremental datasets with SOTA methods. Besides, the visualization shows the effect of prospective transport.
We also conduct ablations to validate the improvement of each part in \mame. We also combine retrospective transport with other methods and report the results in the supplementary.

\subsection{Experiment Settings}
\noindent {\bf Datasets:} Following the protocol defined in ~\cite{rebuffi2017icarl,wu2019large,yu2020semantic}, we evaluate the performance of related methods on CIFAR-100~\cite{krizhevsky2009learning}, CUB200-2011~\cite{WahCUB2002011} and ImageNet ILSVRC2012~\cite{ILSVRC15}. We additionally conduct experiments on a real-world multimedia facial expression recognition task, \ie, RAF-DB~\cite{li2018reliable}.  They are listed as:

{ \bfname{CIFAR-100}}: contains 50,000  training and 10,000 testing images, with a total of 100 classes.
\bfname{CUB-200 and CUB-100}:  a fine-grained image dataset with 200 bird species, including 11,788 images. We also randomly sample 100 out of 200 classes from CUB to form CUB-100, according to~\cite{yu2020semantic}.
{\bfname{ImageNet-1000 and ImageNet-100}}: ImageNet is a large scale dataset with 1,000 classes, with about 1.28 million images for training and 50,000 for validation. We also randomly select 100 classes from the original ImageNet-1000 to form ImageNet-100 according to~\cite{wu2019large}.
{\bfname{RAF-DB}}: comprises 15,339 real-world facial images annotated with one of the seven expression classes. Following ~\cite{zhu2020iexpressnet,kacem2017novel,yang2018facial}, we select 6 basic expressions (without neutral) as the experimental data.

According to the common setting of class-incremental learning~\cite{rebuffi2017icarl}, 
\emph{all the datasets are shuffled with NumPy random seed 1993.\footnote{We follow this benchmark setting for a fair comparison.}}
For the subset datasets, \ie, CUB-100 and ImageNet-100, the sub-sampled classes are the first 100 classes after class shuffle~\cite{wu2019large,yu2020semantic}.
There are two types of incremental setting~\cite{rebuffi2017icarl,yu2020semantic}. The first setting starts from half of the total classes, and makes the rest come in different phases~\cite{hou2019learning,yu2020semantic}, while the other setting fixes the number of classes in the first task the same as  later tasks~\cite{rebuffi2017icarl,zhao2020maintaining}.  In this paper, we conduct experiments with these two settings, and validate the universal performance improvement of \mame.

\noindent {\bf Compared methods:} In this section, we compare \name with the SOTA methods, including iCaRL~\cite{rebuffi2017icarl}, BiC~\cite{wu2019large}, WA~\cite{zhao2020maintaining}.  We also report the offline model, \ie, Oracle, in the results.  

{ \bfname{Finetune}}: finetunes the incremental model with cross-entropy. Finetune does not consider overcoming forgetting, and faces the forgetting phenomena.
\bfname{iCaRL}~\cite{rebuffi2017icarl}: utilizes nearest center mean as classifier, and applies knowledge distillation~\cite{hinton2015distilling} to prevent forgetting. The loss function of iCaRL corresponds to Eq.~\ref{eq:icarl}.
{\bfname{BiC}~\cite{wu2019large}}: trains an extra bias correction layer to remove the bias of linear layer. BiC separates a validation set from exemplars, and the validation set is not used for training.
{\bfname{WA}~\cite{zhao2020maintaining}}: normalizes the fc-layer with ${\ell}_2$ norm, and the layer would not become imbalanced when learning new classes. 
{\bfname{Oracle}}: jointly trains all classes in an offline manner, which can be viewed as the upper bound of CIL methods.

{\noindent \bf Implementation details:} All models are implemented with Pytorch~\cite{paszke2019pytorch}. For CIFAR100, we adopt a 32-layer ResNet~\cite{he2015residual} and train 160 epochs. The learning rate starts from $0.1$, and suffers a decay of $0.1$ at $80$ and $120$ epochs. We adopt an 18-layer ResNet  for ImageNet, CUB, and RAF-DB, training 90 epochs in total. The learning rate begins at $0.1$ and suffers a decay of $0.1$ every $30$ epochs. The models are optimized by SGD with batch size $128$, and the temperature $\tau$ is set to 2. To solve OT problem, we use Sinkhorn algorithm~\cite{cuturi2013sinkhorn,sinkhorn1967concerning} and set the entropic regularization term $\alpha$ to $0.45$.

\subsection{Comparison with SOTA Methods}

We first report results by making the number of classes equal for every task. For CIFAR-100, the 100 classes are shuffled and divided into 2,5,10, and 20  incremental tasks. For ImageNet and CUB, the total classes are divided into 10 incremental tasks. Since all compared methods are exemplar-based, we fix an equal number of exemplars for every method, \ie, 2,000 exemplars for CIFAR-100 and ImageNet-100, 20,000 for ImageNet-1000. As a result, the picked exemplars per class is 20, which is \emph{abundant} for every class.
Correspondingly, we also conduct the experiment on CUB-100/200 with \emph{rare} exemplars, \ie, we only save three exemplars per class. The exemplars are selected by herding algorithm~\cite{welling2009herding}.

The performance curves are shown in Figure~\ref{figure:sota}, and the averaged accuracy is shown in Table~\ref{table:inc}. We report top-5 accuracy for ImageNet and top-1 accuracy for CIFAR and CUB. 
We can infer from the  results that our proposed \name outperforms the current SOTA methods in terms of the final incremental accuracy the average incremental accuracy. 
The trend of results is consistent for the three datasets except for ImageNet100. 
BiC works well with abundant exemplars, especially the ImageNet-100 dataset. However, it needs to build a validation set to tune the extra layer. For CUB-100/200 with rare exemplars, only one exemplar per class can be reserved for validation, and the performance of BiC suffers a decay for easily overfitting. However, \name outperforms it by 20$\%$ with rare exemplars.
The experiment results indicate that our proposed method can well handle incremental learning in both small-scale images and large-scale images, with both abundant and rare exemplars. 

\begin{table}[t] 
	\centering{
		\caption{ \small CIL performance  on CIFAR-100 with 2, 5, 10 and 20   steps. The average results over all the incremental steps  are reported.}
		\vspace{-3mm}
		{\begin{tabular}{l|c|c|c|c}
				\addlinespace
				\toprule
				{\# Incremental steps} &{2}  &{5}  &{10} &{20}    
				\\
				\midrule
				Finetune &  56.34 & 37.98   & 25.81 &16.49   \\
				iCaRL~\cite{rebuffi2017icarl} &  66.05& 61.64     & 61.74 &60.41 \\
				BiC~\cite{wu2019large} &  69.16 & 65.09    &  62.79   &59.12  \\
				WA~\cite{zhao2020maintaining} &  67.56 & 64.91    &  57.43  &55.71 \\
				
				\name &\bf 69.64 &\bf 	68.26 & \bf 	65.48 &\bf 62.98 \\
				\midrule
				Oracle & \multicolumn{3}{|c}{70.1}\\
				\bottomrule
			\end{tabular}\label{table:inc}
	}}
\end{table}

\subsection{Experiments with Vast Base Classes}
In a real-world application such as product categorization or face recognition, incremental learning usually starts from a model trained on a pre-collected dataset. To mimic this, we start from a model trained on half of the total classes~\cite{hou2019learning,yu2020semantic}, and the rest classes come in different phases.
For CIFAR-100, there are 100 classes in total, and we make 50 of them as the base classes in the first task, and make the rest 50 classes emerge in 2, 5, and 10 tasks separately.  

The results are shown in Figure~\ref{figure:sotacifar}, and the averaged accuracy performance is reported in the supplementary. We can infer from the figures that our proposed \name outperforms the current SOTA methods in terms of the final incremental accuracy the average incremental accuracy. 
Additionally, since prospective transport utilizes the semantic relationship between old and new classes to initialize a new classifier,  the performance of the generated new classifier is related to the relationship between old and new classes. As the base classes become more, we can extract more semantic relationships from the old classes that facilitate the learning of new classes, and expect stronger performance of prospective transport. Correspondingly, we find that the performance improvement than the runner-up method in Figure~\ref{figure:sotacifar} is slightly larger than the CIFAR tasks in Figure~\ref{figure:sota}, and such improvement is consistent with our assumption.

\subsection{Visualization of Transported Classifier} \label{sec:vis}

\begin{figure}[t]
	\begin{center}
		\subfigure[ Decision boundary for 3 old classes.]
		{\includegraphics[width=.48\columnwidth]{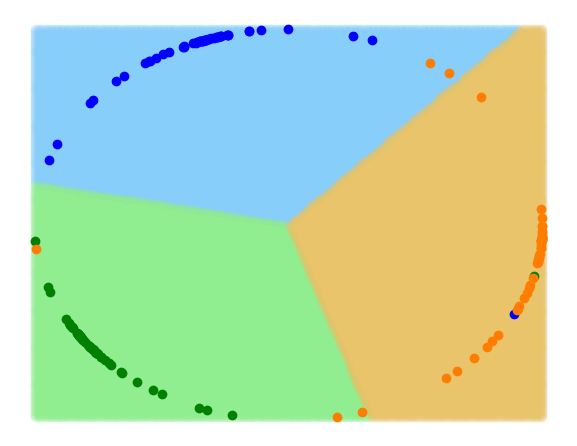} \label{figure:vis1}}
		\subfigure[  PT for 2 new classes.]
		{\includegraphics[width=.48\columnwidth]{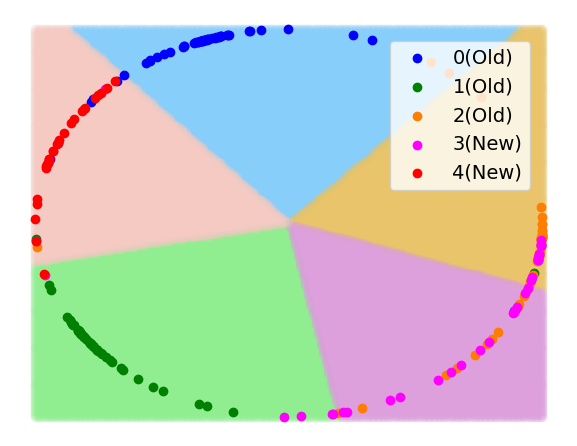}\label{figure:vis2}}
		\vspace{-3mm}
	\end{center}
	\caption{ \small Visualizations of the decision boundary. Dots stand for instances, and shadow regions stand for classification boundaries. We first train with three old classes, and then use optimal transport to initialize the classifier for two new classes (\ie, prospective transport). The initialized  classifier well captures the semantic relationship, and can classify new classes without training them.} \label{figure:visulization}	
	\vspace{-5mm}
\end{figure}

In this part, we visualize the learned decision boundaries on CIFAR-100 dataset. Instances are shown in 2D by learning embedding module $\phi(\cdot):\mathbb{R}^{D} \rightarrow \mathbb{R}^{2}$, \ie, we attach an extra linear layer  to the CNNs as embedding module. Note that we adopt the cosine classifier, and the visualized features are normalized. In the first task, we train a classifier for three classes (\emph{road, palm tree, and snake}). In the second task, two new classes emerge (\emph{bicycle and cloud}). We then utilize the classifier transportation algorithm in Alg.~\ref{alg:ot} to transport the classifier for old classes $W_{old}$ into the new ones $W_{new}$ as the initialization. We show the decision boundary of the augmented classifier over five classes (including old and new classes).

We plot the visualization in Figure~\ref{figure:visulization}. 
In each figure, dots denote instances, and the shadow region represents the model's classification boundary. 
We can infer that the optimal transported classifier well captures the class-wise relationship, and can depict and tell apart old and new classes even not train on them. It also shows the powerful ability of prospective transport, and the transferred classifier can act as a good initialization of new classes, avoiding the negative influence of random initialization. Additionally, in this trial, the old classes are almost unrelated to the new ones, and the initialized classifier can also depict the difference between old and new classes, which works better than the random initialization. We can infer that the initialized classifier would be stronger when old and new classes have a stronger correlation.

\subsection{Real-World Facial Expression Recognition}

In real-world facial expression recognition problems, the expression classes are becoming increasingly fine-grained and incremental.
We also conduct experiments with a real-world multimedia dataset, \ie, RAF-DB. RAF-DB is a facial expression recognition dataset with six classes, and we divide them into two and three tasks to form an incremental stream. Similar to~\cite{zhu2020iexpressnet}, we resize the facial images into 100 $\times$ 100 pixels, and maintain 60 exemplars for each class.

\begin{figure}[t]
	\begin{center}
		\subfigure[ RAF-DB, 3 tasks]
		{\includegraphics[width=.48\columnwidth]{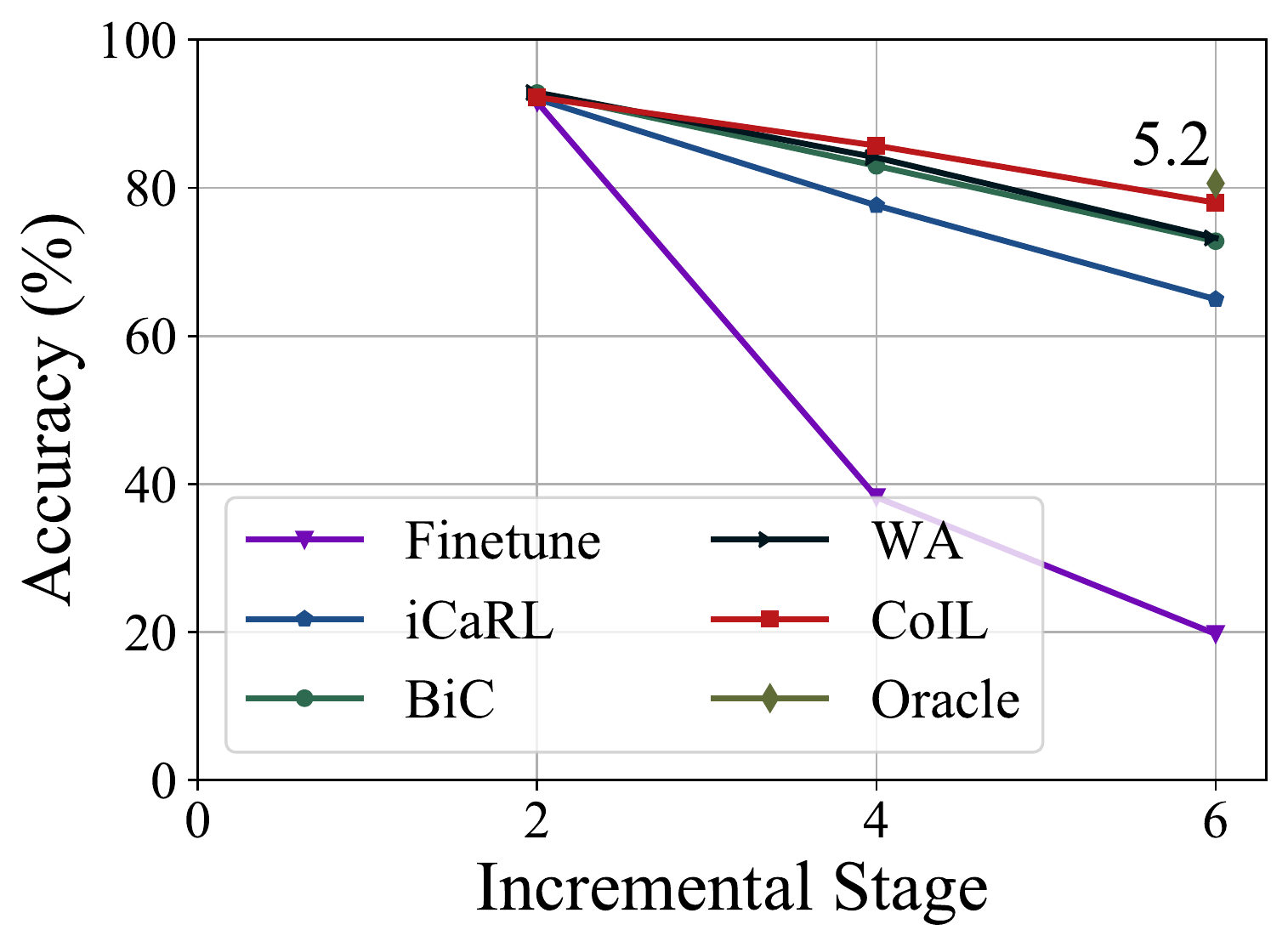} \label{figure:ab1}}
		\subfigure[ RAF-DB, 2 tasks]
		{\includegraphics[width=.48\columnwidth]{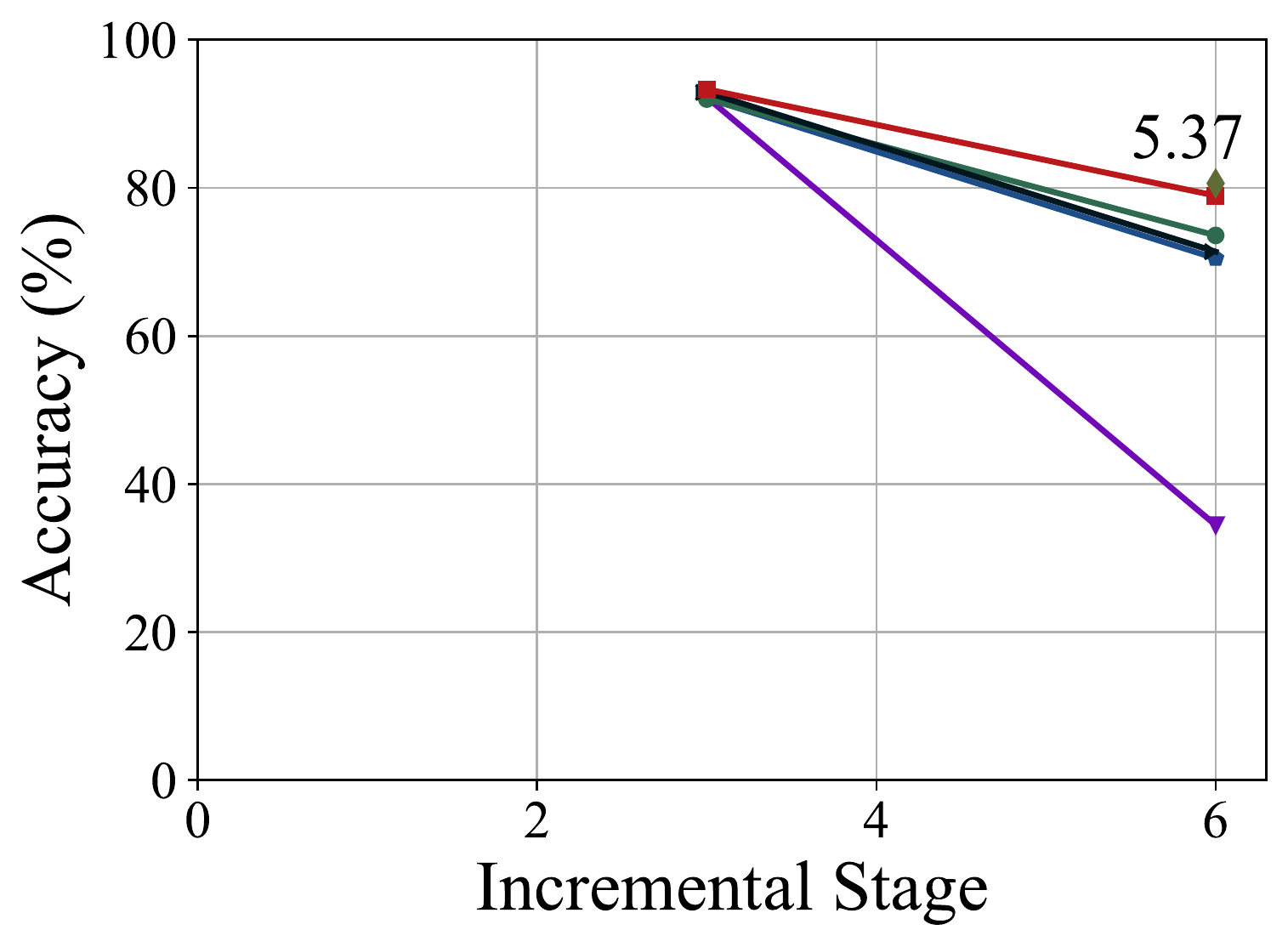}\label{figure:ab2}}
	\end{center}
	\vspace{-3mm}
	\caption{\small Results on facial expression recognition dataset RAF-DB.  } \label{figure:rafdbresults}
	\vspace{-3mm}
\end{figure}

The performance curves are shown in Figure~\ref{figure:rafdbresults}.
Finetuning learns the new facial images without the restriction of former learned knowledge, and the performance drastically decays as new classes are incorporated. iCaRL utilizes the power of knowledge distillation to map the new model with the old model, and preserves the learned knowledge from catastrophic forgetting. WA and BiC extend iCaRL with more restrictions, \ie, weight normalization and extra bias correction layer, and the results are better than vanilla iCaRL. 
However, facial expressions are strongly correlated, and COIL is good at utilizing such relationship, and utilizes the optimal transported knowledge to obtain the best performance. The results validate that under the real-world incremental learning scenarios, \name can still work competitively.

\subsection{Ablation Study}

In this section, we provide an ablation study of the components in \mame. More ablations are reported in supplementary. 

\noindent {\bf Impact of prospective transport:} In addition to the visualization of  decision boundaries,
we also conduct experiences to measure the help of prospective transport to model adaptation quantitatively. 
To evaluate this, we compare different strategy to extend the classifier: {\bf PT:} initialize $W_{new}=T{W_{old}}$, where $T$ is obtained by solving the OT problem in Eq.~\ref{eq:ot}; {\bf NCM:} drop the linear classifier, and classify the new instances by nearest center mean~\cite{rebuffi2017icarl}; {\bf Random:} randomly initialize $W_{new}$. We test on CIFAR-100 with ten tasks.

With these three strategies, when facing a new incremental task, we \emph{directly} test on new classes and report the accuracy in Figure~\ref{figure:ab1}. Compared to the random classifier, which obtains $10\%$ accuracy on the new classes, NCM constructs class means of new classes, and assigns labels via the nearest class center, and performs much better. However, since the embedding module is not fitted to the new classes, NCM cannot extract the most suitable class means. By contrast, PT utilizes the class-wise relationship to construct optimal transportation  mapping, and initialize a good classifier with the old one, which performs best. We also notice that the test performance increases as more known classes, which is consistent with our awareness that the more classes we know, the easier to seek a related class and transfer.

\noindent {\bf Ablation of semantic transport:}  we conduct experiments to validate the effectiveness of each part in \mame. Four variations are included in the comparison: {\bf Variation 1:} training with cross-entropy (Eq.~\ref{eq:crossentropy}); {\bf Variation 2:} training with cross-entropy and distillation loss (Eq.~\ref{eq:icarl}); {\bf Variation 3:} equipping variation 2 with prospective transport; {\bf Variation 4:} equipping variation 2 with retrospective transport. The experiences are conducted with CIFAR-100 of 5 tasks.

Figure~\ref{figure:ab2} reports the results. Since variation 1 only learns new concepts without the constraint of former knowledge, it quickly forgets the knowledge and suffers catastrophic forgetting. Variation 2 utilizes distillation loss to overcome forgetting, and takes a small step to improve performance. Variation 3 and 4 equips variation 2 with an extra transport step, and both get more improvement than variation 2. Notice that prospective transport works at the beginning of the task, and its effect lies more in the model adaptation. In contrast, retrospective transport works throughout the training process, pushing the model away from forgetting and showing to be stronger than variation 3. However, combing them together, we bring \mame, which works better than every single one. \name outperforms the baseline methods by a substantial margin.

\begin{figure}[t]
	\begin{center}
		\subfigure[ Impact of prospective transport]
		{\includegraphics[width=.48\columnwidth]{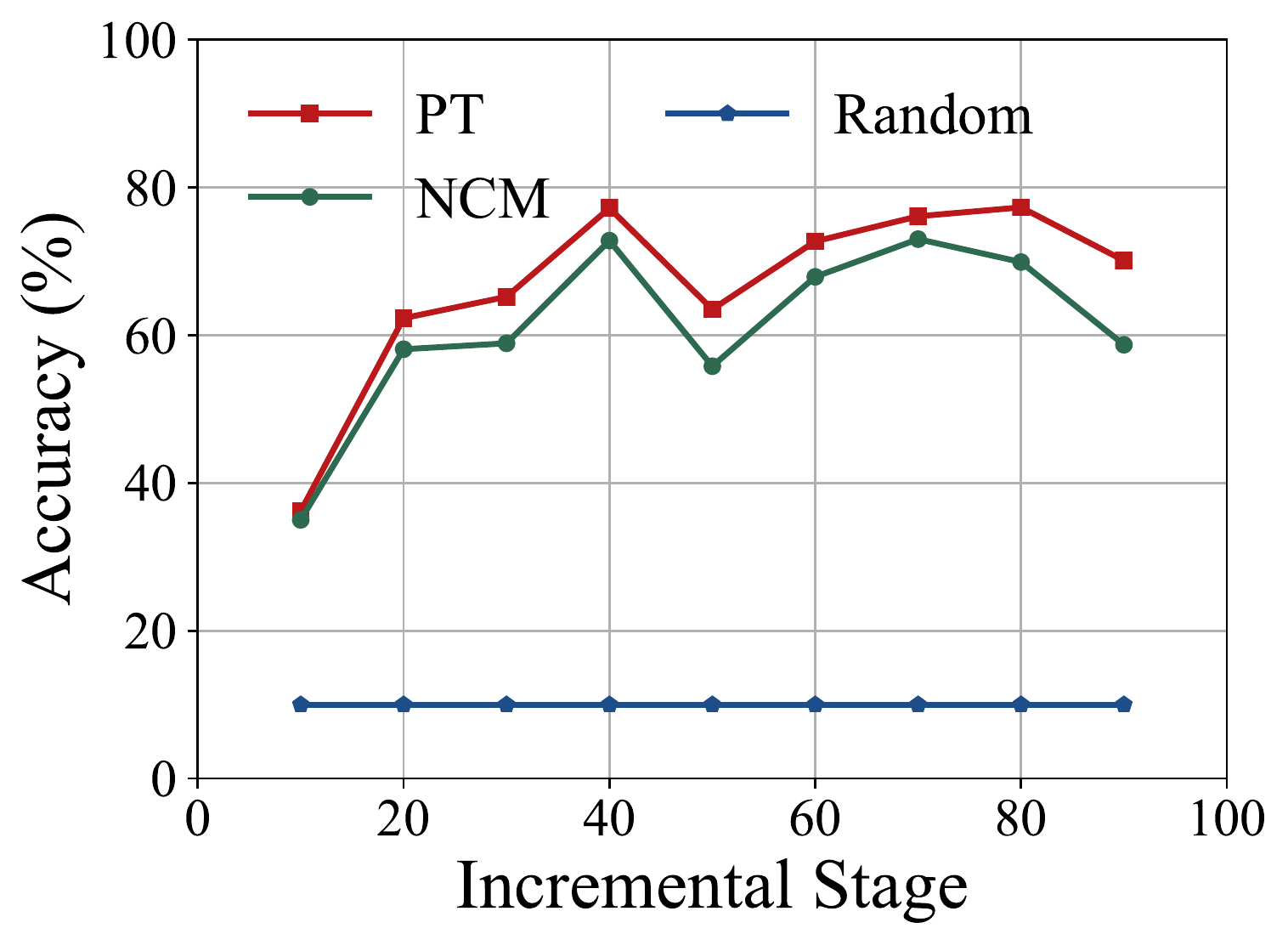} \label{figure:ab1}}
		\subfigure[ Ablation of semantic transport]
		{\includegraphics[width=.48\columnwidth]{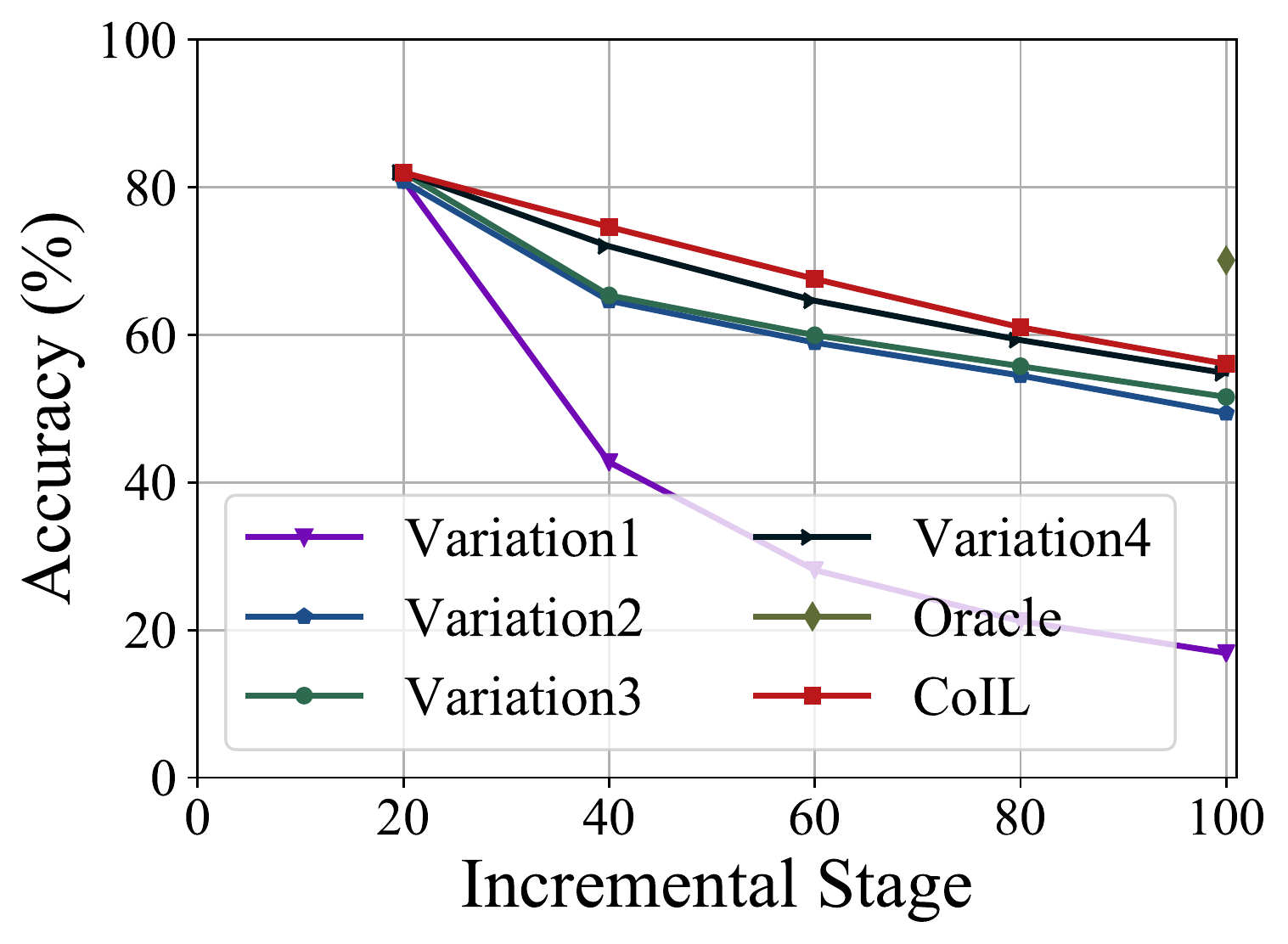}\label{figure:ab2}}
	\end{center}
	\vspace{-3mm}
	\caption{\small Ablation study. Left: semantic-transportation synthesized classifier has better performance than nearest center mean.
		Right:	PT and RT both improve performance than baseline method.} \label{figure:abalation}
	\vspace{-3mm}
\end{figure}

\section{Conclusion}

In real-world applications, learning systems often face instances of new classes. To learn a classifier for all seen classes incrementally without forgetting old classes, class-incremental learning is thus proposed. However, strong relevancy between old and new classes is neglected by current approaches, while we find it can help to facilitate the incremental learning process.
In this paper, we propose \name to utilize the class-wise semantic relationship. On the one hand, 
transporting old class knowledge to new ones helps fast adapting to the new class, and avoids the negative impact of random weight initialization.  
On the other hand, transporting new classifiers as old ones can exert an extra regularization term over the incremental model, which well prevents the catastrophic forgetting phenomena.
The proposed \name efficiently adapts to new classes, and preserves old knowledge when learning new ones. 
How to further explore the marginal probability  and the transport cost between classes are interesting future works.

\section*{Acknowledgments}

This research was supported by National Key
R\&D Program of China (2020AAA0109401), NSFC (61773198, 61921006, 62006112), NSFC-NRF Joint Research Project under Grant 61861146001, Nanjing University Innovation Program for
Ph.D. candidate (CXYJ21-53), Collaborative Innovation Center of Novel Software Technology and Industrialization, NSF of Jiangsu Province (BK20200313).

	\bibliographystyle{ACM-Reference-Format}
\balance
\bibliography{mm21}

\end{document}